\documentclass{article}
\usepackage{iclr2026_conference,times}

\usepackage{amsmath,amsfonts,bm}

\def\eqref#1{equation~\ref{#1}}

\def\1{\bm{1}}

\DeclareMathAlphabet{\mathsfit}{\encodingdefault}{\sfdefault}{m}{sl}
\SetMathAlphabet{\mathsfit}{bold}{\encodingdefault}{\sfdefault}{bx}{n}

\usepackage{hyperref}
\usepackage{url}

\usepackage{graphicx}
\usepackage{subcaption}
\usepackage[export]{adjustbox}
\usepackage{booktabs}
\usepackage{float}
\usepackage{tabularx}
\usepackage{makecell}

\title{Benchmarking Open-Source Layout Detection Models for Data Snapshot Extraction from Institutional Documents}

\author{AJ Carl P. Dy$^{\dagger}$ \& Aivin V. Solatorio$^{*}$ \\
Development Data Group \\
Office of the World Bank Group Chief Statistician \\
The World Bank \\
1818 H Street N.W., \\
Washington, 20433 \\
District of Columbia, USA \\
\texttt{\{ady, asolatorio\}@worldbank.org}
}

\iclrfinalcopy 
\begin{document}

\maketitle

\begingroup
\renewcommand\thefootnote{\fnsymbol{footnote}}
\footnotetext[1]{GitHub/HF: \href{https://github.com/avsolatorio}{\texttt{@avsolatorio}}, \href{mailto:avsolatorio@gmail.com}{avsolatorio@gmail.com}}
\footnotetext[2]{GitHub/HF: \href{https://github.com/ajdajd}{\texttt{@ajdajd}}, \href{mailto:ajcarl.dy@gmail.com}{ajcarl.dy@gmail.com}}
\endgroup

\begin{abstract}

Institutional documents contain substantial amounts of operational and analytical information embedded within figures and tables. Current approaches for extracting visual content from documents are largely built around generic document layout analysis, where figures and tables are treated as uniformly relevant document objects rather than semantically meaningful analytical artifacts. In this work, we introduce a benchmark dataset and evaluation framework for \textit{data snapshot extraction}, the task of identifying and localizing semantically meaningful visual artifacts within institutional documents. The benchmark spans humanitarian reports, World Bank policy research working papers, and project appraisal documents, and includes annotations for figures and tables that contain reusable analytical information. Using this dataset, we benchmarked multiple open-source layout detection models and evaluated both detection performance and spatial extraction quality. Our results show that current models struggle to generalize to operational institutional documents despite strong performance on conventional academic benchmarks. Common failure modes include confusion between analytical and non-analytical content, fragmentation of composite analytical artifacts, and incomplete extraction of contextual information required for interpretation. These findings highlight a persistent gap between generic document layout analysis and operationally useful data snapshot extraction. We release the source PDFs, annotation dataset, metadata, and source code to support future research in operational document intelligence. The dataset is available at https://huggingface.co/datasets/ai4data/data-snapshot and the source code is available at https://github.com/worldbank/ai4data/tree/main/experimental/data-snapshot.

\end{abstract}

\section{Introduction}

Development, humanitarian, and public policy organizations continuously publish institutional documents containing substantial amounts of operational and analytical information~\citep{worldbankpubrecord2010, datafordevelopment2017}. Organizations such as the World Bank, UNHCR, and humanitarian coordination clusters regularly produce reports featuring statistical tables, monitoring visualizations, financing summaries, implementation matrices, and geospatial maps. These visual artifacts often contain rich information that may not be fully discussed in the surrounding text or that extend beyond the narrative context in which they are presented.

AI systems have increasingly become the primary modality for extracting, retrieving, and synthesizing insights from documents~\citep{layoutllm2024}. However, much of the information contained in these visual artifacts remains difficult to access in downstream analytical workflows because it is embedded within figures and tables rather than narrative text. Existing document processing pipelines, including OCR-based systems and retrieval workflows built primarily around extracted text streams, often inadequately capture the semantic structure and contextual completeness of visual content. OCR pipelines may fragment content into disconnected text segments while losing important spatial relationships, titles, legends, and layout context~\citep{Linga2025}. Similarly, large language model workflows may process visual regions inconsistently or inefficiently when they are embedded within long documents~\citep{layoutllm2024}. These limitations can result in significant information loss when operational insights exist primarily inside visual structures, reducing the reliability of downstream tasks such as structured extraction, indexing, summarization, and multimodal reasoning.

Such limitations are particularly important in operational documents where the narrative text summarizes only a subset of the findings while additional analytical information remains embedded within tables and figures. Humanitarian reports typically describe high-level trends in the discussion while detailed incident distributions are present only inside charts and monitoring visuals. Policy research papers discuss conclusions narratively while econometric outputs and comparative statistics remain embedded within tables. Project appraisal documents describe interventions in the text while financing allocations, procurement structures, and implementation details are only given as structured matrices. Actual examples from our benchmark corpora are provided in \autoref{sec:samples}.

To address this challenge, we formalize the extraction of semantically meaningful visual artifacts from institutional documents as a distinct document understanding task. This approach allows us to tackle the problem in a principled manner and systematically assess existing solutions.

\subsection{Defining data snapshots}

Our approach builds on the broader field of document layout analysis, which is concerned with the identification and structural understanding of visual elements within documents, such as text blocks, figures, tables, headers, and other layout components~\citep{led2026, doclaynet2022}. Existing layout detection systems and benchmarks are primarily designed for generic document segmentation and object localization, treating figures and tables as uniformly relevant extraction targets regardless of their analytical utility~\citep{doclaynet2022, vgtd4la2023}. In practice, institutional documents frequently contain decorative photographs, logos, cover images, tables of contents, recommendation blocks, and formatting tables, which are technically classified as figures or tables, but do not contain reusable analytical information.

In this work, we focus on a narrower but operationally important subset of visual artifacts that contain semantically meaningful analytical content. We refer to the extraction of these regions from documents as \textbf{data snapshot extraction}. A \textbf{data snapshot} is defined as a bounded visual region within a document page containing structured or semi-structured information intended for analytical interpretation or operational reuse. It represents a semantically meaningful analytical artifact and may include contextual elements such as titles, legends, captions, footnotes, or adjacent visual components when they are necessary for interpretation. Examples include analytical charts, statistical tables, financing matrices, dashboard-style monitoring summaries, and geospatial visualizations. Representative examples of both valid and invalid data snapshots from our benchmark corpora are provided in \autoref{sec:samples}. This distinction creates what we refer to as the \textbf{data snapshot extraction gap}: the inability of current document AI systems to reliably identify and isolate semantically meaningful visual regions containing reusable operational information.

\subsection{Why specialized snapshot detection still matters}
\label{sec:why-specialized-snapshot-detection-still-matters}

The data snapshot extraction gap persists even as recent multimodal large language models demonstrate increasingly strong capabilities in understanding figures and tables directly from page images~\citep{led2026}. While these systems can reason over visual content end-to-end, applying general-purpose multimodal pipelines to large-scale institutional document collections remains computationally expensive~\citep{doclayoutyolo2024}.

In most institutional documents, only a small fraction of content contains analytically relevant visual information. Across our benchmark corpora, data snapshots occupy a median of only 31.3\% of page area on pages where they are present. Furthermore, analysis of a representative subset of the benchmark indicates that data snapshots typically appear on only about one in five pages within a document.\footnote{As an illustrative example, a typical document page rendered at 300 DPI (2550$\times$3300 pixels) requires approximately 765 tokens under the image token accounting scheme published for GPT-4o-class vision models. Processing pages only containing data snapshots therefore has the potential to substantially reduce multimodal inference costs.} As a result, pipelines that process entire documents uniformly may spend substantial computational resources analyzing irrelevant content.

This creates a strong incentive for accurate and efficient snapshot localization systems that can isolate data snapshots prior to downstream processing. Such systems can reduce both the amount of visual content requiring expensive multimodal processing and the amount of irrelevant context propagated into downstream pipelines.

\subsection{Benchmarking existing open-source systems}

The data snapshot extraction gap also raises an important practical question: how well do existing open-source document layout detection systems generalize to operational institutional documents? While recent layout analysis models demonstrate strong performance on established benchmarks~\citep{doclaynet2022,doclayoutyolo2024}, their effectiveness on institutional documents remains unclear. To investigate this question, we present a benchmark for evaluating data snapshot extraction from institutional documents spanning humanitarian reports, policy research working papers, and project appraisal documents. The benchmark emphasizes real-world operational document complexity, including dense infographic layouts, noisy visual structures, and analytically irrelevant visual objects.

Rather than proposing a new model architecture, our objective is to establish realistic baselines for current open-source capabilities. We benchmarked multiple off-the-shelf document layout detection systems in order to evaluate how well existing models generalize to operational institutional documents. 

This work makes four primary contributions. First, we constructed a benchmark dataset for semantically meaningful data snapshot extraction spanning humanitarian reports, policy research papers, and project appraisal documents. Second, we developed an evaluation framework that jointly evaluates object detection quality and spatial extraction quality. Third, we benchmarked multiple open-source layout detection systems on our dataset. Finally, we release the source PDFs, annotations, metadata, and source code to support future research.

\section{Related work}

\subsection{Document layout analysis}

Document layout analysis aims to identify and understand the structural organization of documents by detecting elements such as text blocks, titles, figures, tables, lists, and other page components~\citep{led2026}. Early work in document analysis focused on rule-based and OCR-centric approaches, while recent advances have increasingly leveraged deep learning, transformer-based architectures, and multimodal document encoders~\citep{vgtd4la2023}. The availability of large-scale benchmarks such as PubLayNet~\citep{publaynet2019}, DocLayNet~\citep{doclaynet2022}, TableBank~\citep{tablebank2020}, and D4LA~\citep{vgtd4la2023} has substantially accelerated progress in document parsing, layout understanding, and information extraction. These datasets provide annotations for document elements, enabling supervised training and evaluation of document parsing systems. Together, these benchmarks have facilitated rapid advances in document layout analysis and document understanding. Models such as LayoutLM~\citep{layoutlm2020} and its successors integrate textual, visual, and spatial information for document understanding, while more recent architectures such as DocLayout-YOLO~\citep{doclayoutyolo2024} emphasize efficient page-level localization across heterogeneous document collections. These systems have found applications in document digitization, intelligent search, OCR enhancement, document retrieval, information extraction, and downstream document understanding pipelines.

Despite these advances, existing benchmarks and evaluation protocols remain focused on the geometric localization and semantic classification of document objects such as figures and tables. Layout elements are typically evaluated according to their geometric and categorical correctness, without regard to whether the detected regions contain analytically meaningful information for downstream use. Consequently, current document layout systems are optimized to identify document structures rather than to distinguish between analytically useful visual artifacts and visually similar but operationally irrelevant content.

\subsection{Operational document intelligence}

Recent interest in document intelligence has expanded beyond scientific publishing and business documents to include development, humanitarian, and public-sector information systems~\citep{ai4data2024, mlwb2023}. Existing efforts in these domains have largely concentrated on text extraction, information retrieval, entity extraction, and knowledge discovery~\citep{beyondkeywords2024, ai4data2024}. Comparatively less attention has been given to the problem of identifying and isolating semantically meaningful visual regions that contain reusable analytical information~\citep{humanitariannlp2023, infoextractionhumanitarian2024}. Consequently, there remains limited understanding of how well contemporary layout detection systems generalize from conventional document benchmarks to operational institutional documents~\citep{vgtd4la2023, doclaynet2022}.

\section{Evaluated models}

We evaluated four open-source document layout detection models spanning transformer-based document encoders and YOLO-based document detectors. These models were selected to provide coverage across both transformer-based and YOLO-based document layout detection approaches while remaining fully reproducible using publicly available model weights.

\paragraph{TF-ID-Large}
TF-ID-Large~\citep{TF-ID} is a vision-language document layout detection model built by fine-tuning Florence-2~\citep{florence22023} on the TF-ID arXiv papers dataset containing manually annotated tables and figures from academic papers. We included TF-ID-Large as a representative transformer-based approach to document layout detection with strong reported performance on scientific document benchmarks. We used the model variant that was trained on the dataset that explicitly annotate figure and table captions.

\paragraph{DocLayout-YOLO}
DocLayout-YOLO~\citep{doclayoutyolo2024} extends the YOLOv10 architecture for document understanding tasks. The model is pretrained on the synthetic DocSynth-300K corpus and fine-tuned on downstream layout-analysis datasets including DocLayNet~\citep{doclaynet2022}, D4LA~\citep{vgtd4la2023}. We included DocLayout-YOLO as a representative high-performance YOLO-based document detector specifically optimized for layout analysis. We used the variant fine-tuned on the developers' own curated dataset, DocStructBench, which we found to be the best for our use-case.

\paragraph{YOLOv11 for Advanced Document Layout Analysis}
YOLOv11 for Advanced Document Layout Analysis~\citep{yolo11} is a document layout detector built on the Ultralytics YOLO11 architecture and fine-tuned on the DocLayNet benchmark dataset. This model represents a lightweight and operationally efficient document layout detection baseline. We used the medium variant for all experiments.

\paragraph{YOLOv26 for Advanced Document Layout Analysis}
YOLOv26 for Advanced Document Layout Analysis~\citep{yolo26} is based on the newer Ultralytics YOLO26 architecture and trained on the updated DocLayNet v1.2 dataset. We included this model as a representative implementation of a recent YOLO-based document layout detector trained on a contemporary dataset. We used the medium variant for all experiments.

\section{The benchmark}

\subsection{Motivation}

During early exploration, we surveyed existing datasets, layout detection models, OCR pipelines, and multimodal document models. While existing models demonstrated strong performance on conventional layout benchmarks, we found that they frequently failed to distinguish between semantically meaningful data snapshots and visually similar but operationally irrelevant layout objects. For example, a figure containing economic indicators and a decorative photograph are typically assigned the same semantic label, while a statistical table and a table of contents are often treated as equivalent tabular structures despite their fundamentally different informational value. These observations revealed a lack of evaluation resources capable of quantifying how well models identify analytically useful visual content, motivating the construction of a dedicated benchmark for data snapshot extraction.

\subsection{Corpus composition}

The benchmark combines three institutional document corpora spanning humanitarian reporting, academic policy research, and development project operations.

\paragraph{UNHCR / ReliefWeb}
This corpus includes humanitarian and protection-related reports containing monitoring and displacement summaries, incident visualizations, protection monitoring summaries, and geospatial maps. Documents were retrieved from the ReliefWeb portal\footnote{https://reliefweb.int} and restricted to analysis documents published by the United Nations High Commissioner for Refugees (UNHCR).

\paragraph{Policy Research Working Papers (PRWP)}
This corpus includes World Bank Policy Research Working Papers containing econometric tables, comparative statistics, analytical figures, and dense academic layouts. Documents were retrieved from the World Bank Documents \& Reports repository\footnote{https://documents.worldbank.org/en/publication/documents-reports\label{foot:wbrepo}} and filtered to Policy Research Working Papers.

\paragraph{Refugee Project Appraisal Documents (PADs)}
This corpus includes refugee-related project appraisal documents containing financing matrices, procurement tables, implementation summaries, institutional diagrams, and results frameworks. Documents were retrieved from the World Bank Documents \& Reports repository\footref{foot:wbrepo} and filtered to English-language Project Appraisal Documents containing refugee-related projects.

Together, these corpora expose models to substantially different document structures, visual conventions, and reporting styles, enabling evaluation across a diverse range of institutional document formats. Table~\ref{tab:dataset_stats} summarizes the composition of each corpus.

\begin{table}[htbp]
    \centering
    \caption{Dataset composition across institutional document corpora.}
    \label{tab:dataset_stats}
    \begin{tabular}{lrrrrr}
        \toprule
        \textbf{Corpus} & \textbf{PDFs} & \textbf{Pages} & \textbf{Figures} & \textbf{Tables} & \textbf{\makecell{Data snapshots \\ (Figures + Tables)} } \\
        \midrule
        UNHCR / ReliefWeb & 338 & 2,765 & 2,198 & 282 & 2,480 \\
        PRWP & 100 & 3,023 & 339 & 590 & 929 \\
        Refugee PADs & 38 & 1,929 & 62 & 437 & 499 \\
        \midrule
        Total & 476 & 7,717 & 2,599 & 1,309 & 3,908 \\
        \bottomrule
    \end{tabular}
\end{table}

\subsection{Sampling and dataset characteristics}

Different sampling strategies were employed across corpora. For the UNHCR corpus, documents were first filtered to reports containing at most 15 pages. TF-ID-Large was then used as a preliminary screening tool to maximize annotation efficiency, where only documents with detected figures or tables on at least 30\% of pages were retained for annotation. This procedure was intended solely to prioritize documents likely to contain analytical visual content and did not affect the subsequent annotation process. In contrast, documents in the PRWP and Refugee PAD corpora were filtered to at most 40 pages and 60 pages, respectively, and then sampled at random without additional content-based filtering.

\begin{figure}[htbp]
    \centering
    \includegraphics[width=0.85\linewidth]{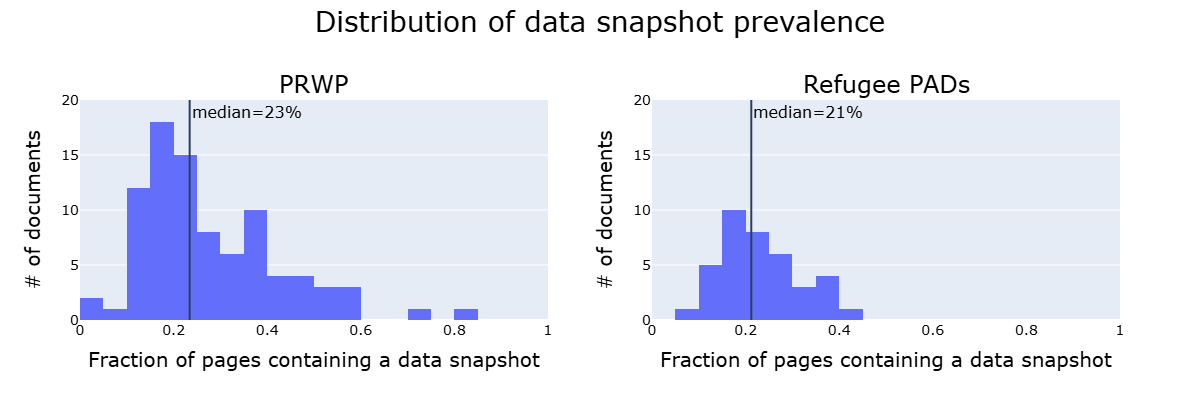}
    \caption{Distribution of the fraction of pages containing at least one data snapshot within the PRWP and Refugee PAD corpora. Median prevalence is approximately one page in five for both corpora. The UNHCR corpus is excluded because documents were screened for analytical visual content during sampling.}
    \label{fig:snapshot-prevalence}
\end{figure}

Because the UNHCR corpus was intentionally enriched for documents likely to contain analytical visual content, statistics describing the prevalence of data snapshots within documents are reported only for the PRWP and Refugee PAD corpora. Figure~\ref{fig:snapshot-prevalence} shows the distribution of the fraction of pages containing at least one data snapshot across these two. Consistent with the observations discussed in Section~\ref{sec:why-specialized-snapshot-detection-still-matters}, data snapshots typically appear on only a minority of pages within most documents, with a median prevalence of approximately one page in five.

\subsection{Annotation workflow}

Annotations were produced using a semi-assisted human-in-the-loop workflow. DocLayout-YOLO was used to generate preliminary figure annotations, while YOLOv11 was used to generate preliminary table annotations. These pre-labels served solely to accelerate the annotation process and were not used directly as ground truth. Instead, all annotations were manually reviewed and corrected page-by-page using Label Studio~\citep{labelstudio}, with every final bounding box verified by the authors. This workflow ensured that the released benchmark reflects human-validated labels rather than model predictions while improving annotation efficiency.

\subsection{Data availability}

To support reproducible research, we release the original PDFs, document metadata, bounding box annotations, benchmarking code, model adapters, and utility scripts. The dataset is publicly available on Hugging Face, while the benchmarking framework and source code are released on GitHub.

\section{Evaluation framework}

\subsection{Detection evaluation}

We evaluate data snapshot extraction as a bounding box detection task over two classes: \textit{Figure} and \textit{Table}. Following standard object detection evaluation protocols~\citep{voc2010, coco2014}, predicted bounding boxes are matched against ground-truth annotations using Intersection-over-Union (IoU). A prediction is considered a true positive if its IoU with a ground-truth object exceeds a threshold of 0.5. Matching is performed using greedy one-to-one assignment, ensuring that each prediction and each ground-truth object can participate in at most one match.

Based on the resulting matches, we calculate and report Precision and Recall. Precision measures the fraction of predicted snapshots that correspond to valid ground-truth objects, while Recall measures the fraction of ground-truth snapshots successfully detected by the model. Together, these two metrics characterize a model's ability to identify semantically meaningful data snapshots while avoiding irrelevant detections.

We intentionally report results using a fixed threshold of IoU = 0.5 rather than COCO-style mean Average Precision (mAP) averaged across multiple IoU thresholds~\citep{coco2014}. Our objective is not generic object detection benchmarking but to evaluate whether a model can reliably identify semantically meaningful data snapshots within institutional documents. In practice, the more important question is whether a model successfully identifies a usable data snapshot rather than how detection performance varies across many IoU thresholds. A fixed IoU threshold therefore provides a simple and interpretable measure of detection success while remaining consistent with common object detection evaluation practice.

\subsection{Spatial extraction quality}

Object detection metrics alone are insufficient for evaluating data snapshot extraction. A prediction may technically satisfy IoU = 0.5 while still excluding critical information such as figure titles, table headers, chart legends, subtitles, footnotes, or axes labels. While detection metrics evaluate whether a snapshot was successfully identified, spatial metrics evaluate how accurately its semantic extent was captured.

\begin{figure}[htbp]
    \centering
    \begin{subfigure}[b]{0.42\textwidth}
        \centering
        \includegraphics[max width=\linewidth, max height=0.20\textheight, keepaspectratio, clip]{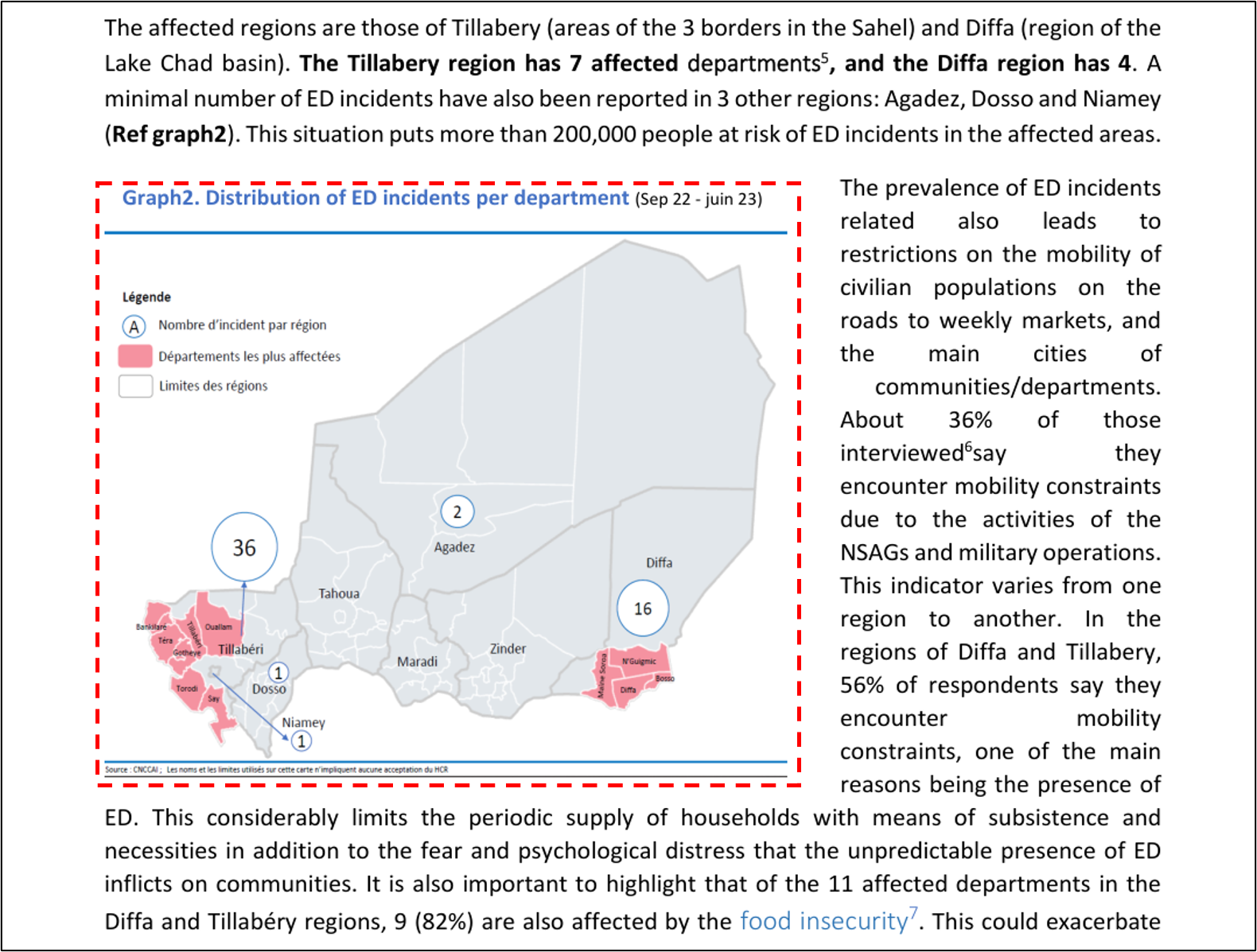}
        \caption{}
    \end{subfigure}
    \hspace{0.02\textwidth}
    \begin{subfigure}[b]{0.42\textwidth}
        \centering
        \includegraphics[max width=\linewidth, max height=0.20\textheight, keepaspectratio, clip]{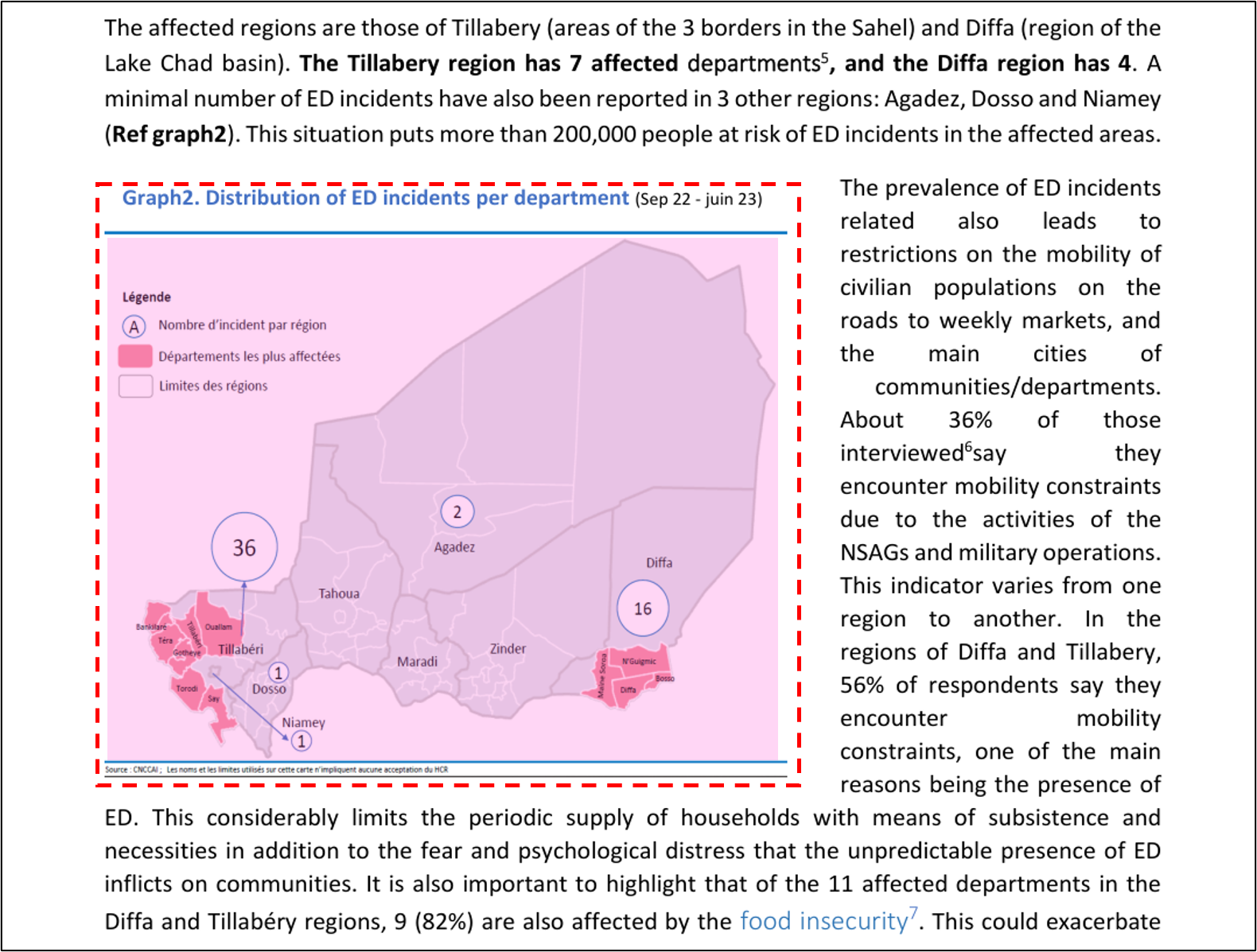}
        \caption{}
    \end{subfigure}
    
    \vspace{0.5em}
    \begin{subfigure}[b]{0.42\textwidth}
        \centering
        \includegraphics[max width=\linewidth, max height=0.20\textheight, keepaspectratio, clip]{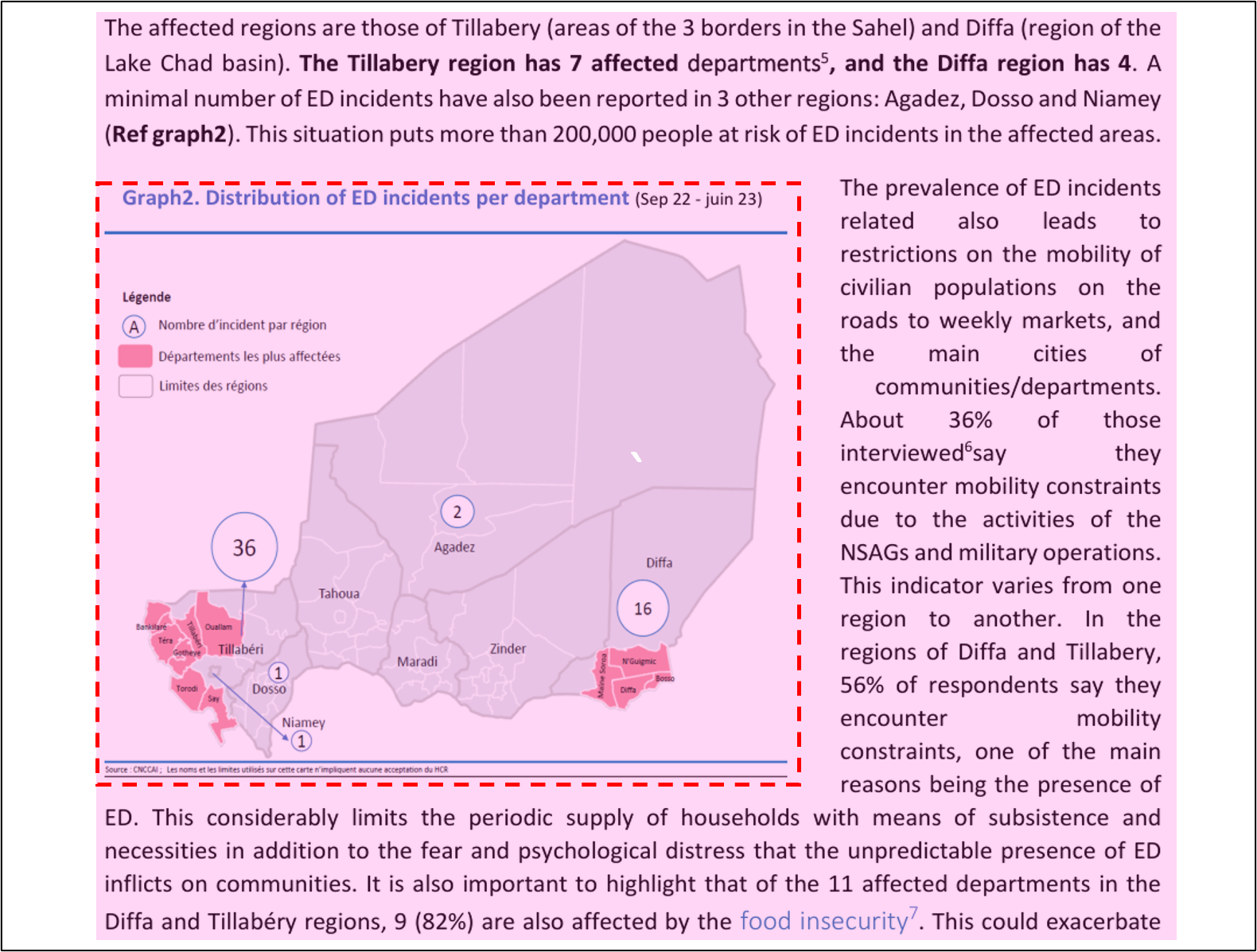}
        \caption{}
    \end{subfigure}
    \hspace{0.02\textwidth}
    \begin{subfigure}[b]{0.42\textwidth}
        \centering
        \includegraphics[max width=\linewidth, max height=0.20\textheight, keepaspectratio, clip]{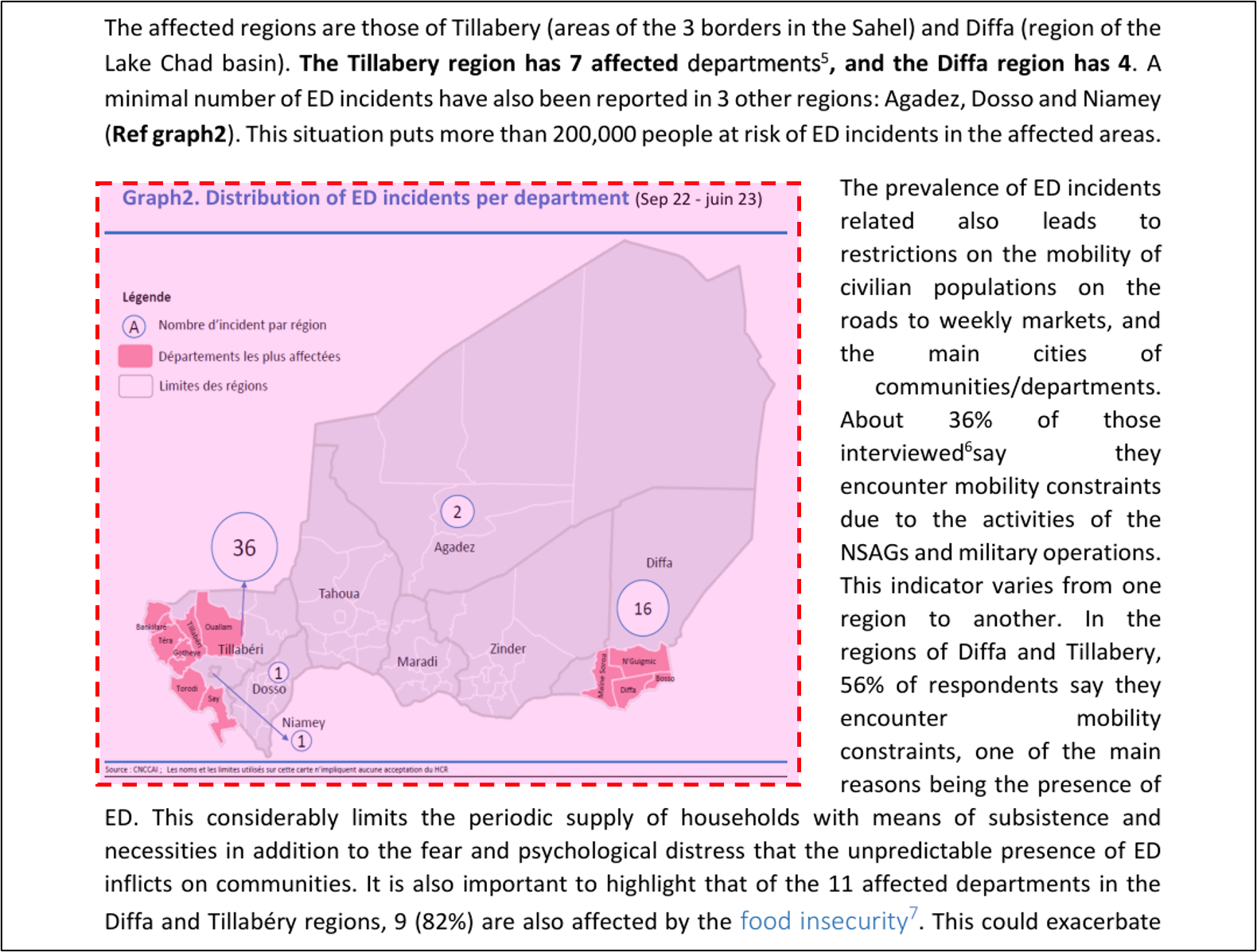}
        \caption{}
    \end{subfigure}

    \caption{Illustration of spatial extraction quality. The dashed rectangle indicates the annotated snapshot extent.
    (a) Reference document region containing a data snapshot.
    (b) Underextraction, where only the main figure is captured while contextual elements such as the title and source footnote are omitted, resulting in low Area Recall.
    (c) Overextraction, where substantial surrounding text and unrelated content are included, resulting in low Area Precision.
    (d) Extraction aligned with the annotated snapshot extent, correctly including the figure, title, and source information, resulting in high Area Recall, high Area Precision, and high IoU.}
    \label{fig:spatial-metrics}
\end{figure}

To capture the completeness and purity of the extracted snapshots, we separately evaluate spatial extraction quality using three complementary metrics: Area Recall, Area Precision, and Intersection-over-Union. Figure~\ref{fig:spatial-metrics} illustrates the relationship between the annotated snapshot extent and several possible extraction outcomes. Note that these spatial metrics are computed only for matched prediction--ground-truth pairs.

\paragraph{Area Recall (Coverage)}
Area Recall measures the fraction of the ground-truth region captured by the predicted bounding box. High coverage is important because downstream extraction pipelines require semantically complete crops. Missing titles, legends, or explanatory footnotes may significantly reduce the usefulness of the extracted snapshot.

\begin{equation}
Area\ Recall = \frac{|A_{pred} \cap A_{gt}|}{|A_{gt}|}
\end{equation}

\paragraph{Area Precision (Purity)}
Area Precision measures the fraction of the predicted bounding box belonging to the ground-truth object. High purity indicates that the extracted crop contains minimal irrelevant surrounding content. 

\begin{equation}
Area\ Precision = \frac{|A_{pred} \cap A_{gt}|}{|A_{pred}|}
\end{equation}

\paragraph{Intersection-over-Union (IoU)}
Intersection-over-Union (IoU) provides a single metric that jointly captures both properties of Area Recall and Area Precision. It quantifies the degree of spatial agreement between a predicted bounding box and its corresponding ground-truth annotation.

\begin{equation}
IoU = \frac{|A_{pred} \cap A_{gt}|}{|A_{pred} \cup A_{gt}|}
\end{equation}

In this benchmark, IoU is used both for matching predictions to ground-truth objects and for evaluating the spatial alignment quality of matched pairs.

\subsection{Bounding box filtering}
\label{subsec:bbox-filtering}

During benchmarking, we applied a post-processing filter that removed predicted bounding boxes with a normalized area smaller than 0.008. Such detections are typically associated with logos, icons, and other small graphical elements rather than meaningful data snapshots. Preliminary inspection showed that this filtering increased precision while producing negligible changes in recall. The filter was applied uniformly across all evaluated models and corpora.

\section{Results}

Tables~\ref{tab:metrics-overall-figure} and~\ref{tab:metrics-overall-table} summarize overall detection and spatial extraction performance for figures and tables, respectively. Tables~\ref{tab:metrics-split-figure} and~\ref{tab:metrics-split-table} provide the same metrics broken down by corpus. All reported results use the bounding-box filtering procedure described in Section~\ref{subsec:bbox-filtering}.

Across both figures and tables, a consistent tradeoff emerges between detection performance and spatial extraction quality. TF-ID-Large generally achieves the highest scores across the spatial extraction metrics, while the YOLO-based models achieve substantially higher detection recall. Among the YOLO-based approaches, DocLayout-YOLO consistently outperforms YOLOv11 and YOLOv26 across most metrics and corpora.

\begin{table}[htbp]
    \centering
    \caption{Figure detection performance and extraction quality across all documents}
    \label{tab:metrics-overall-figure}
    \begin{tabularx}{\textwidth}{ll *{4}{>{\centering\arraybackslash}X}}
        \toprule
        Category & Metric & TF-ID-Large & DocLayout-YOLO & YOLOv11 medium & YOLOv26 medium \\
        \midrule
        Detection        & Precision      & \textbf{0.628} & 0.547 & 0.378 & 0.388 \\
                         & Recall         & 0.488 & \textbf{0.802} & 0.761 & 0.721 \\
        Spatial Accuracy & IoU            & \textbf{0.877} & 0.820 & 0.817 & 0.817 \\
                         & Area Precision & 0.935 & \textbf{0.996} & 0.990 & 0.990 \\
                         & Area Recall    & \textbf{0.938} & 0.824 & 0.826 & 0.826 \\
        \bottomrule
    \end{tabularx}
\end{table}

\begin{table}[htbp]
    \centering
    \caption{Table detection performance and extraction quality across all documents}
    \label{tab:metrics-overall-table}
    \begin{tabularx}{\textwidth}{ll *{4}{>{\centering\arraybackslash}X}}
        \toprule
        Category & Metric & TF-ID-Large & DocLayout-YOLO & YOLOv11 medium & YOLOv26 medium \\
        \midrule
        Detection        & Precision      & \textbf{0.562} & 0.468 & 0.487 & 0.460 \\
                         & Recall         & 0.861 & \textbf{0.893} & 0.862 & 0.891 \\
        Spatial Accuracy & IoU            & \textbf{0.919} & 0.834 & 0.817 & 0.824 \\
                         & Area Precision & 0.972 & 0.993 & \textbf{0.994} & 0.994 \\
                         & Area Recall    & \textbf{0.946} & 0.841 & 0.822 & 0.829 \\
        \bottomrule
    \end{tabularx}
\end{table}

\begin{table}[htbp]
    \centering
    \caption{Figure detection performance and extraction quality, split by corpus}
    \label{tab:metrics-split-figure}
    \begin{tabularx}{\textwidth}{ll *{4}{>{\centering\arraybackslash}X}}
        \toprule
        Category & Metric & TF-ID-Large & DocLayout-YOLO & YOLOv11 medium & YOLOv26 medium \\
        \midrule
        \multicolumn{6}{l}{\textbf{UNHCR / ReliefWeb}} \\
        \midrule
        Detection                         & Precision      & \textbf{0.614} & 0.549 & 0.399 & 0.398 \\
                                          & Recall         & 0.435 & \textbf{0.794} & 0.758 & 0.709 \\
        Spatial Accuracy                  & IoU            & \textbf{0.867} & 0.822 & 0.819 & 0.818 \\
                                          & Area Precision & 0.922 & \textbf{0.995} & 0.988 & 0.989 \\
                                          & Area Recall    & \textbf{0.941} & 0.826 & 0.829 & 0.829 \\
        \midrule
        \multicolumn{6}{l}{\textbf{PRWP}} \\
        \midrule
        Detection                        & Precision      & \textbf{0.737} & 0.607 & 0.431 & 0.430 \\ 
                                         & Recall         & 0.802 & \textbf{0.829} & 0.794 & \textbf{0.829} \\ 
        Spatial Accuracy                 & IoU            & \textbf{0.904} & 0.794 & 0.804 & 0.806 \\ 
                                         & Area Precision & 0.976 & \textbf{0.998} & \textbf{0.998} & 0.996 \\ 
                                         & Area Recall    & \textbf{0.925} & 0.796 & 0.806 & 0.810 \\ 
        \midrule
        \multicolumn{6}{l}{\textbf{Refugee PADs}} \\
        \midrule
        Detection                        & Precision      & \textbf{0.432} & 0.331 & 0.098 & 0.128 \\ 
                                         & Recall         & 0.661 & \textbf{0.919} & 0.694 & 0.548 \\ 
        Spatial Accuracy                 & IoU            & \textbf{0.919} & 0.885 & 0.814 & 0.851 \\ 
                                         & Area Precision & 0.964 & 0.998 & \textbf{0.998} & 0.985 \\ 
                                         & Area Recall    & \textbf{0.954} & 0.887 & 0.815 & 0.865 \\ 
        \bottomrule
    \end{tabularx}
\end{table}

\begin{table}[htbp]
    \centering
    \caption{Table detection performance and extraction quality, split by corpus}
    \label{tab:metrics-split-table}
    \begin{tabularx}{\textwidth}{ll *{4}{>{\centering\arraybackslash}X}}
        \toprule
        Category & Metric & TF-ID-Large & DocLayout-YOLO & YOLOv11 medium & YOLOv26 medium \\
        \midrule
        \multicolumn{6}{l}{\textbf{UNHCR / ReliefWeb}} \\
        \midrule
        Detection                        & Precision      & 0.460 & \textbf{0.510} & 0.462 & 0.408 \\
                                         & Recall         & 0.826 & \textbf{0.947} & 0.936 & 0.936 \\
        Spatial Accuracy                 & IoU            & \textbf{0.913} & 0.855 & 0.838 & 0.849 \\
                                         & Area Precision & 0.969 & 0.997 & 0.994 & \textbf{0.999} \\
                                         & Area Recall    & \textbf{0.942} & 0.858 & 0.844 & 0.850 \\
        \midrule
        \multicolumn{6}{l}{\textbf{PRWP}} \\
        \midrule
        Detection                        & Precision      & \textbf{0.797} & 0.689 & 0.698 & 0.694 \\
                                         & Recall         & \textbf{0.941} & 0.914 & 0.902 & 0.924 \\
        Spatial Accuracy                 & IoU            & \textbf{0.942} & 0.812 & 0.802 & 0.802 \\ 
                                         & Area Precision & 0.982 & \textbf{0.998} & 0.996 & 0.997 \\ 
                                         & Area Recall    & \textbf{0.960} & 0.813 & 0.805 & 0.804 \\ 
        \midrule
        \multicolumn{6}{l}{\textbf{Refugee PADs}} \\
        \midrule
        Detection                        & Precision      & \textbf{0.423} & 0.305 & 0.337 & 0.324 \\
                                         & Recall         & 0.776 & \textbf{0.831} & 0.760 & 0.817 \\
        Spatial Accuracy                 & IoU            & \textbf{0.886} & 0.852 & 0.825 & 0.839 \\ 
                                         & Area Precision & 0.959 & 0.981 & \textbf{0.992} & 0.985 \\ 
                                         & Area Recall    & \textbf{0.926} & 0.870 & 0.832 & 0.852 \\ 
        \bottomrule
    \end{tabularx}
\end{table}

\subsection{Qualitative examples}

\begin{figure}[htbp]
    \centering
    \begin{subfigure}[b]{0.42\textwidth}
        \centering
        \includegraphics[max width=\linewidth, max height=0.20\textheight, keepaspectratio, clip]{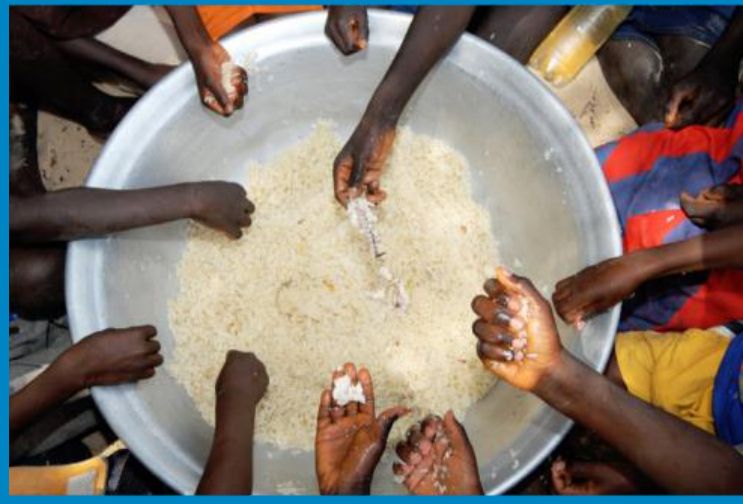}
        \caption{}
    \end{subfigure}
    \hspace{0.02\textwidth}
    \begin{subfigure}[b]{0.42\textwidth}
        \centering
        \includegraphics[max width=\linewidth, max height=0.20\textheight, keepaspectratio, clip]{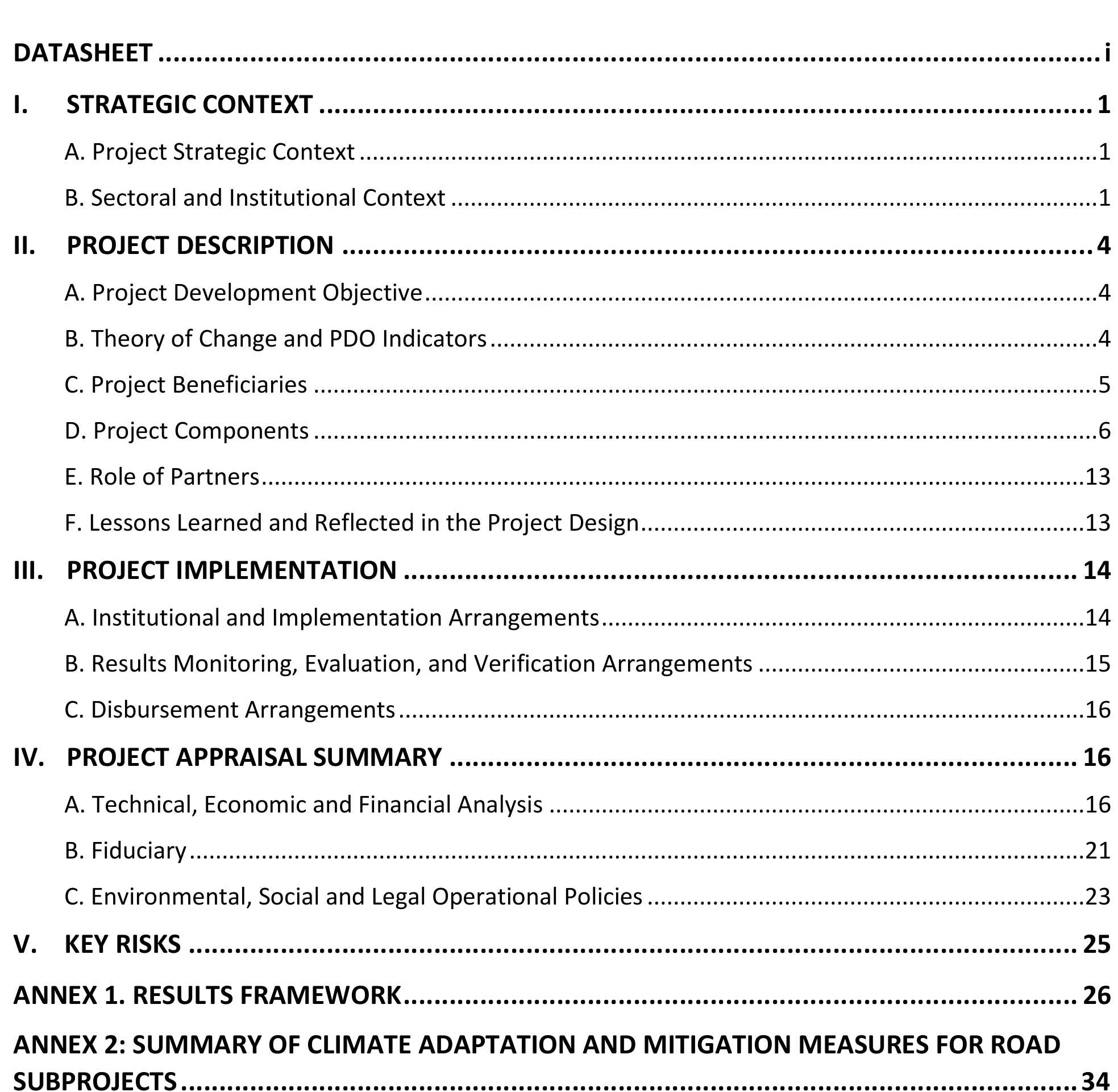}
        \caption{}
    \end{subfigure}
    
    \vspace{0.5em}
    \begin{subfigure}[b]{0.42\textwidth}
        \centering
        \includegraphics[max width=\linewidth, max height=0.20\textheight, keepaspectratio, clip]{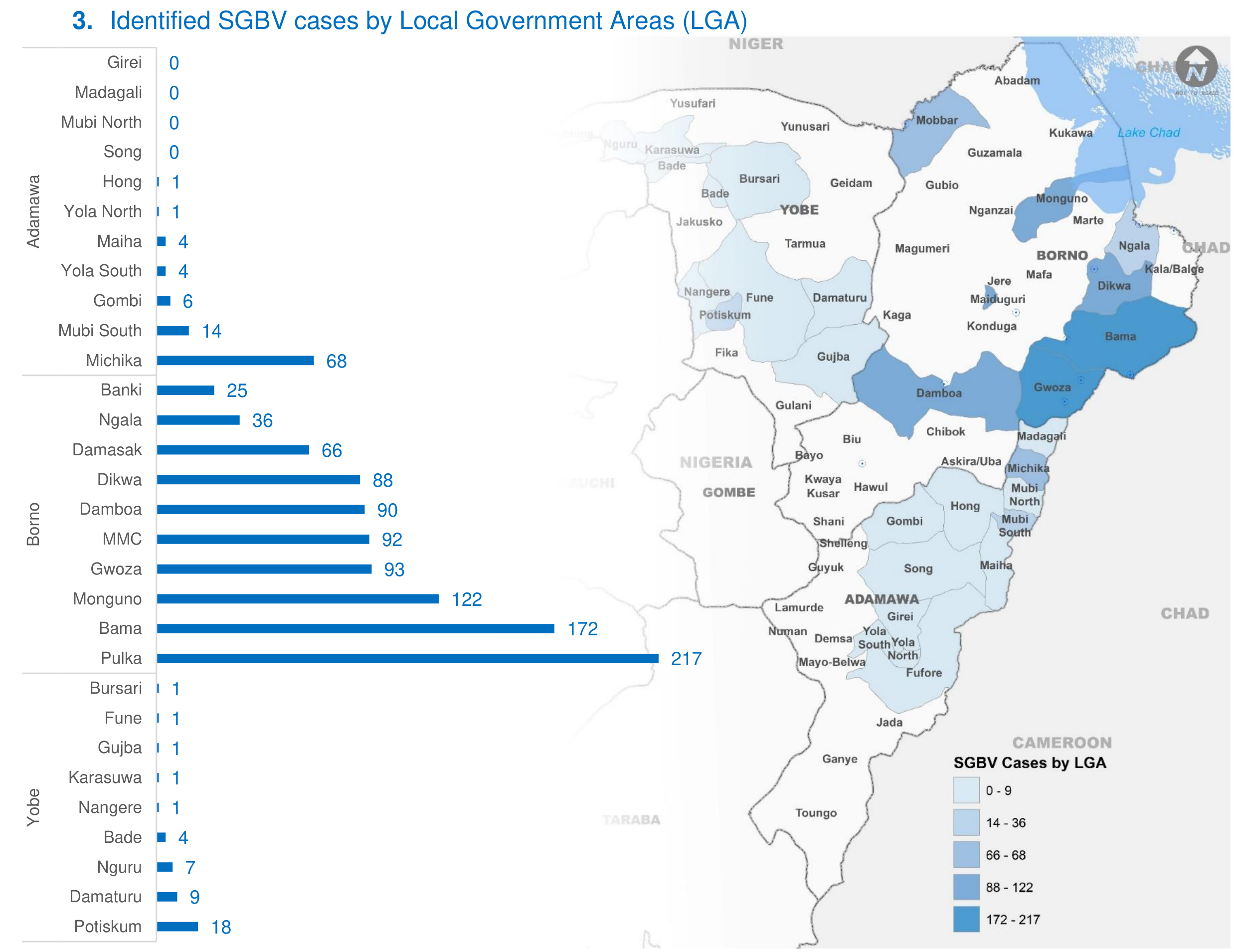}
        \caption{}
    \end{subfigure}
    \hspace{0.02\textwidth}
    \begin{subfigure}[b]{0.42\textwidth}
        \centering
        \includegraphics[max width=\linewidth, max height=0.20\textheight, keepaspectratio, clip]{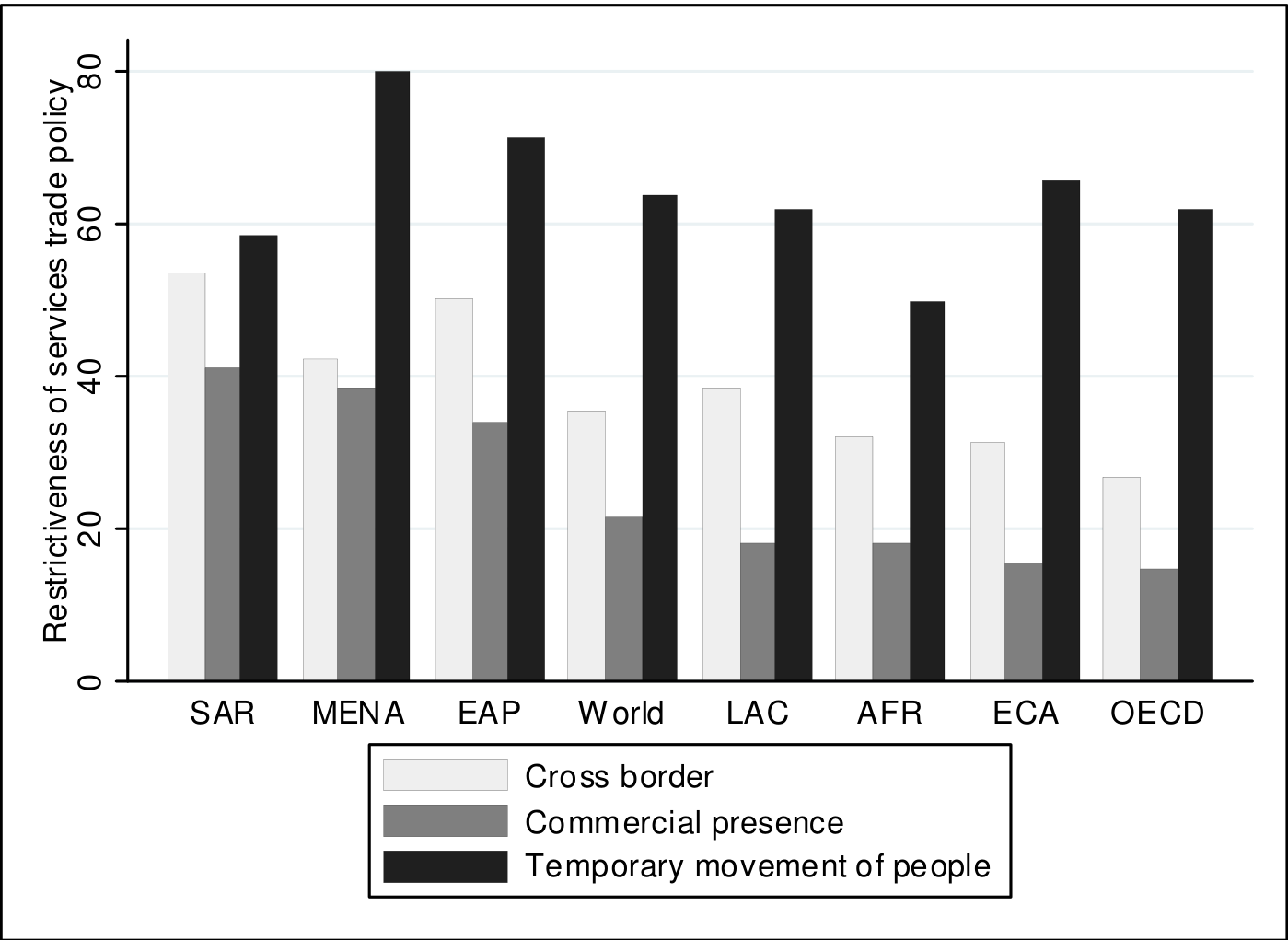}
        \caption{}
    \end{subfigure}

    \caption{Representative failure modes observed across evaluated models.
    (a) Decorative humanitarian photograph incorrectly extracted as a figure.
    (b) Table of contents incorrectly extracted as a table.
    (c) Operational dashboard missed entirely.
    (d) Partial extraction excluding title and footnotes despite satisfying IoU=0.5.}
    \label{fig:failure-modes}
\end{figure}

While aggregate metrics provide a quantitative view of model performance, qualitative inspection reveals recurring failure modes that are not fully captured by summary statistics. Figure~\ref{fig:failure-modes} presents representative examples observed across the benchmark corpora. These examples illustrate several of the challenges discussed in Section~\ref{sec:analysis-and-findings}, including confusion between analytical and non-analytical content, fragmentation of composite analytical artifacts, missed detections in visually complex layouts, and incomplete extractions that omit contextual elements required for interpretation.

\section{Analysis and findings}
\label{sec:analysis-and-findings}

Our experiments reveal several recurring patterns across models and document domains, summarized in Tables~\ref{tab:metrics-overall-figure}--\ref{tab:metrics-split-table}.

First, the primary challenge in data snapshot extraction is not merely detecting figures and tables, but distinguishing analytically meaningful artifacts from visually similar but operationally irrelevant content. Across all models, figure precision remains relatively modest, ranging from 0.378--0.628 despite substantially higher recall values of 0.488--0.802 as shown in Table~\ref{tab:metrics-overall-figure}. Manual inspection shows that many false positives correspond to valid document objects under conventional layout analysis, including logos, decorative images, photographs, organizational charts, and flowcharts. Similar patterns are observed for tables, where common false positives include tables of contents, recommendation tables, abbreviation lookups, and administrative formatting tables. These results suggest that existing layout detectors remain optimized for generic figure and table identification rather than semantic differentiation of analytical content. In this sense, the benchmark is measuring a more demanding task than conventional layout detection, where many of these objects would be considered correct predictions.

Second, composite visual artifacts remain particularly challenging. Humanitarian reports frequently contain dashboard-style monitoring summaries, infographics, statistical factoids, data cards, and other composite layouts composed of multiple charts, maps, and explanatory panels intended to be interpreted together. During annotation, these artifacts were treated as single data snapshots because their analytical meaning emerges from the combination of their constituent elements. However, models frequently fragmented these composites into multiple detections corresponding to individual subcomponents. This behavior was especially common in the UNHCR corpus, where a substantial fraction of annotated figures consist of composite analytical artifacts rather than standalone charts or maps. These observations suggest that current layout models are effective at identifying individual visual elements but struggle to infer the semantic boundaries of larger analytical artifacts.

Third, figures are consistently more challenging than tables. As shown in Tables~\ref{tab:metrics-split-figure} and~\ref{tab:metrics-split-table}, table recall remains high across all evaluated models, ranging from 0.760--0.947 across corpora, while figure recall exhibits substantially greater variability. This pattern suggests that current models generalize more reliably to tabular structures than to visual artifacts. While tables generally follow a relatively constrained set of visual conventions, institutional figures span a much broader range of layouts, compositions, and presentation styles. As a result, figures present a substantially more diverse recognition problem than tables.

Fourth, training data appears to strongly influence both snapshot boundary definition and cross-domain generalization. As shown in Tables~\ref{tab:metrics-overall-figure} and~\ref{tab:metrics-overall-table}, TF-ID-Large consistently achieves the strongest spatial extraction quality, attaining the highest figure IoU (0.877) and Area Recall (0.938), as well as the highest table IoU (0.919) and Area Recall (0.946). Manual inspection suggests that this advantage is due to the inclusion of contextual elements such as titles, subtitles, footnotes, and nearby explanatory text that are omitted by the YOLO-based models. This behavior is consistent with its training corpus, which contains academic figures and tables accompanied by standardized captions. In some cases, TF-ID-Large extends detections (albeit sometimes incorrectly) beyond the core visual object to include text resembling academic captions. Existing layout datasets often treat titles, captions, footnotes, and figures as separate objects, whereas the data snapshot formulation treats these elements as part of a single analytical artifact when they are necessary for interpretation. This difference in annotation philosophy may partially explain TF-ID-Large's stronger spatial extraction performance.

A similar training-data effect is evident among the YOLO-based models. DocLayout-YOLO consistently achieves stronger performance than YOLOv11 and YOLOv26 across most metrics and corpora. One possible explanation is that its pretraining and fine-tuning datasets expose the model to a substantially broader range of document structures than the DocLayNet-centered training pipelines used by YOLOv11 and YOLOv26. This diversity more closely resembles the variability encountered in institutional documents and may partially explain its stronger generalization performance.

In summary, the benchmark reveals a persistent domain gap between conventional layout benchmarks and operational institutional documents. Existing layout datasets largely emphasize academic publications and generic document structures, whereas institutional documents often contain visual and semantic characteristics that are underrepresented in conventional benchmarks. The performance degradation observed across the humanitarian corpus suggests the presence of a visual domain gap, where complex analytical artifacts and reporting formats differ substantially from those represented in current training datasets. At the same time, the prevalence of false positives on semantically irrelevant figures and tables highlights a semantic domain gap: conventional layout benchmarks do not require models to distinguish between generic document objects and analytically meaningful content.

These findings highlight a persistent gap between benchmark-oriented layout understanding and operationally useful data snapshot extraction. While current document layout systems can successfully localize many figures and tables, identifying semantically meaningful analytical artifacts and capturing their complete contextual extent remains a substantially more challenging problem.

\section{Discussion}

The results of this study suggest that semantically meaningful data snapshot extraction remains an open problem in document intelligence. Although contemporary layout detection systems demonstrate strong performance on established benchmarks, all evaluated models exhibited substantial limitations when applied to operational institutional documents. Common failure modes included confusion between analytical and non-analytical content, fragmentation of composite analytical artifacts, and incomplete extraction of contextual elements required for interpretation. These limitations suggest that current open-source systems remain insufficient for reliable deployment across large institutional document collections.

Beyond evaluation, the findings highlight the broader importance of data snapshot extraction within modern document intelligence pipelines. Many downstream applications, including retrieval, indexing, structured extraction, summarization, and multimodal reasoning, depend on accurate localization of semantically meaningful visual artifacts. By isolating analytical content prior to expensive multimodal processing, specialized snapshot extraction systems may reduce computational costs, improve retrieval quality, and preserve contextual information required for downstream reasoning. More broadly, the data snapshot formulation shifts the unit of analysis from individual layout objects toward semantically complete analytical artifacts, aligning document understanding more closely with how humans consume operational information.

Finally, the benchmark introduced in this work provides a foundation for systematic research on data snapshot extraction. Unlike existing layout datasets, the benchmark explicitly captures analytical relevance, contextual completeness, and semantic boundaries, providing supervision for models trained specifically for operational document workflows. By releasing the source documents, annotations, metadata, benchmarking code, and evaluation framework, we aim to provide a foundation for future datasets, models, and evaluation methodologies that are better aligned with the requirements of operational document intelligence.

\section{Limitations}

This work has several limitations.

First, the novelty of data snapshots as a distinct analytical artifact within the broader context of document layout analysis introduces a degree of annotation subjectivity that is not present in conventional layout detection tasks. Decisions regarding the semantic boundaries of composite artifacts, dashboard-style summaries, and contextual elements such as titles, captions, and footnotes may admit multiple reasonable interpretations. Although annotation guidelines were developed and consistently applied throughout the benchmark construction process, alternative definitions of snapshot extent are possible.

Second, the benchmark represents data snapshots using rectangular bounding boxes. While this choice is consistent with common document layout evaluation practices, many analytical artifacts exhibit complex structures that do not align perfectly with rectangular bounding boxes. As illustrated by several examples in \autoref{sec:non-rectangular}, stylistic layout choices and dense page designs may prevent some analytical artifacts from being cleanly represented using rectangular bounding boxes. Future versions of the benchmark may benefit from segmentation masks or hierarchical annotations that more accurately capture the internal structure of composite visual artifacts.

Third, the benchmark currently focuses on figures and tables and evaluates detection and spatial extraction quality rather than downstream semantic interpretation. Consequently, the benchmark measures whether semantically meaningful artifacts are successfully localized, but does not directly evaluate the accuracy of information extraction, retrieval, or reasoning performed on the resulting snapshots.

Fourth, the dataset emphasizes development, humanitarian, and public-sector document ecosystems and may not fully represent the diversity of institutional publishing formats encountered in other domains. Additional corpora from government agencies, international organizations, and private-sector institutions may further improve benchmark coverage.

Finally, this work focuses on benchmark construction and evaluation rather than model development. While the results demonstrate substantial room for improvement in current open-source systems, the design and training of specialized data snapshot extraction models remains an important direction for future research.

\section{Conclusion}

This work introduced a benchmark dataset and evaluation framework for data snapshot extraction from institutional documents spanning humanitarian reports, policy research papers, and project appraisal documents. By focusing on semantically meaningful analytical artifacts rather than generic layout objects, the benchmark addresses an important gap between conventional document layout analysis and operational document intelligence workflows.

Our evaluation of multiple open-source layout detection systems demonstrates that strong performance on established layout benchmarks does not necessarily translate to strong performance on operational institutional documents. Across models, common failure modes included confusion between analytical and non-analytical content, fragmentation of composite analytical artifacts, and incomplete extraction of contextual information required for interpretation.

More broadly, our findings suggest that data snapshot extraction constitutes a distinct document understanding task with challenges that are not fully captured by existing benchmarks. We hope that the public release of the dataset, annotations, source documents, and evaluation framework will support future research on models, datasets, and evaluation methodologies for operational document intelligence.

\subsubsection*{Acknowledgments}
This work is supported by the “Artificial intelligence for understanding data use and enhancing knowledge discovery” project funded by the World Bank-UNHC Joint Data Center on Forced Displacement - RA-P503405-RESE-TF0C7575.

\subsubsection*{Disclaimer and disclosure of AI use}
The findings, interpretations, and conclusions expressed in this paper are entirely those of the authors. They do not necessarily represent the views of the International Bank for Reconstruction and Development/World Bank and its affiliated organizations, or those of the Executive Directors of the World Bank or the governments they represent.

This work used AI tools at various stages, including open source AI models for layout detection. In addition, NotebookLM and ChatGPT were employed to enhance the manuscript’s readability. We also thank Rafael Sevilla Macalaba for providing technical advice and suggestions to strengthen the implementation and manuscript.

\bibliographystyle{iclr2026_conference}
\bibliography{references}

\newpage

\appendix

\section{Appendix: Representative data snapshots}
\label{sec:samples}

This appendix provides representative examples of data snapshots from each corpus included in the benchmark. The examples illustrate the diversity of analytical artifacts encountered in operational institutional documents and provide intuition for the distinction between semantically meaningful data snapshots and generic layout objects.

For each example, we briefly describe the information contained within the snapshot, explain why it satisfies the data snapshot definition used in this work, and discuss potential downstream uses of the extracted information.

\subsection{UNHCR / ReliefWeb}

The UNHCR corpus contains the most visually diverse collection of data snapshots in the benchmark. Unlike academic publications, where figures often consist of standalone charts accompanied by captions, humanitarian reports frequently communicate information through composite visual artifacts that combine charts, maps, indicators, explanatory text, and graphical elements within a single analytical unit intended to be interpreted as a whole.

\begin{figure}[htbp]
    \centering
    \begin{subfigure}[b]{0.48\textwidth}
        \centering
        \includegraphics[max width=\linewidth, max height=0.18\textheight, keepaspectratio]{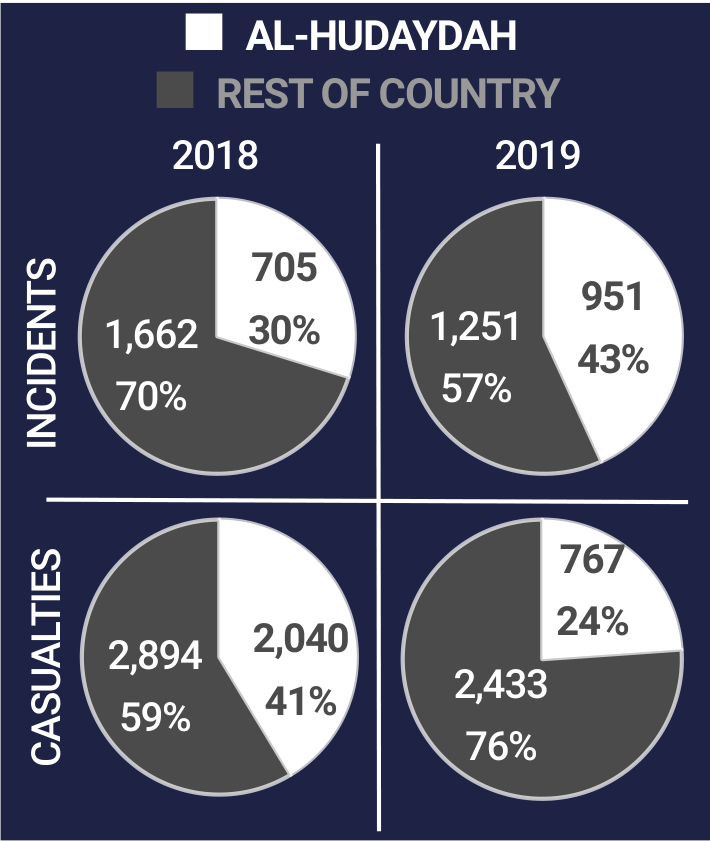}
        \caption{}
        \label{fig:unhcr-a}
    \end{subfigure}
    \hfill
    \begin{subfigure}[b]{0.48\textwidth}
        \centering
        \includegraphics[max width=\linewidth, max height=0.18\textheight, keepaspectratio]{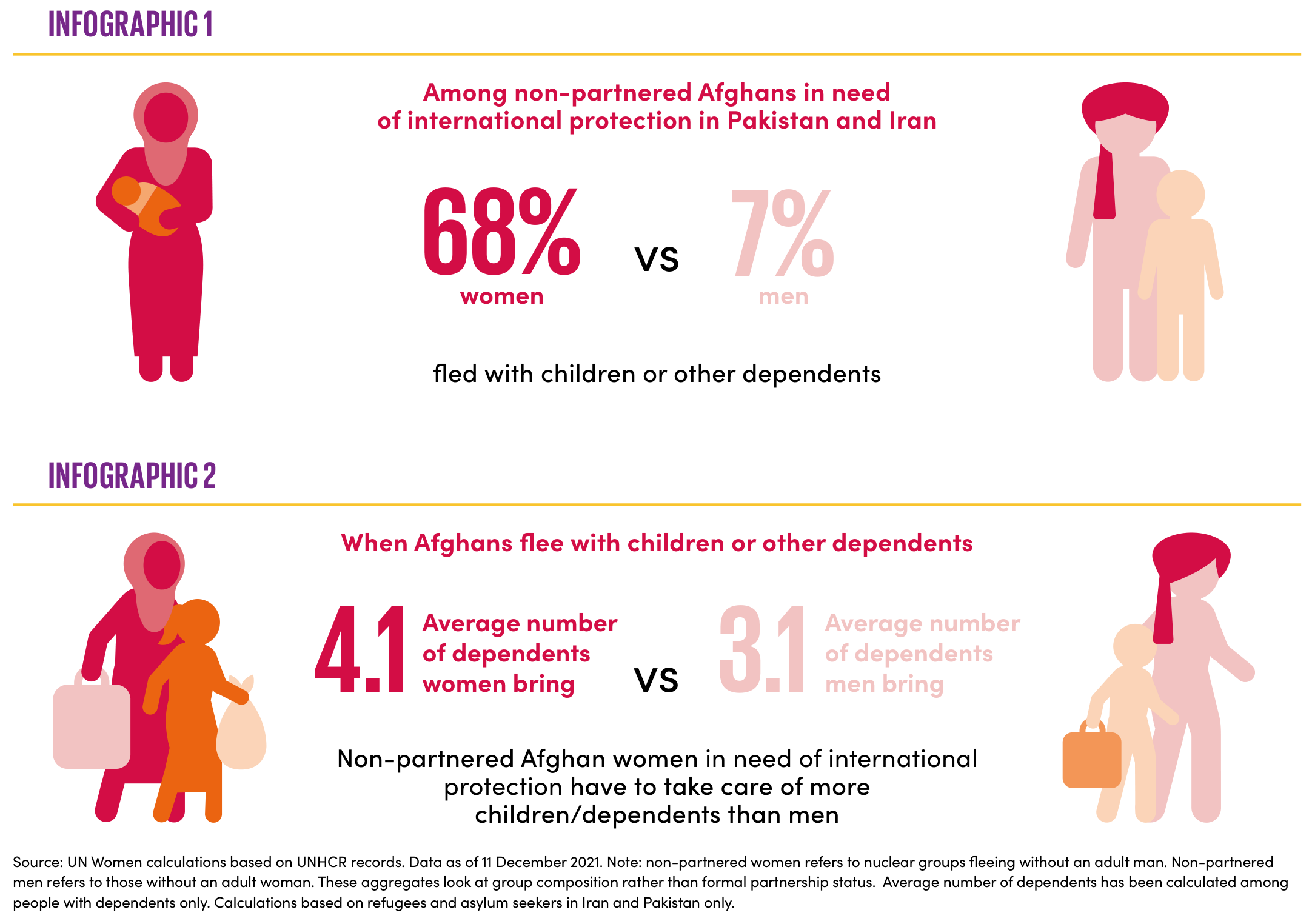}
        \caption{}
        \label{fig:unhcr-b}
    \end{subfigure}
    
    \vspace{0.5em}
    \begin{subfigure}[b]{0.48\textwidth}
        \centering
        \includegraphics[max width=\linewidth, max height=0.18\textheight, keepaspectratio]{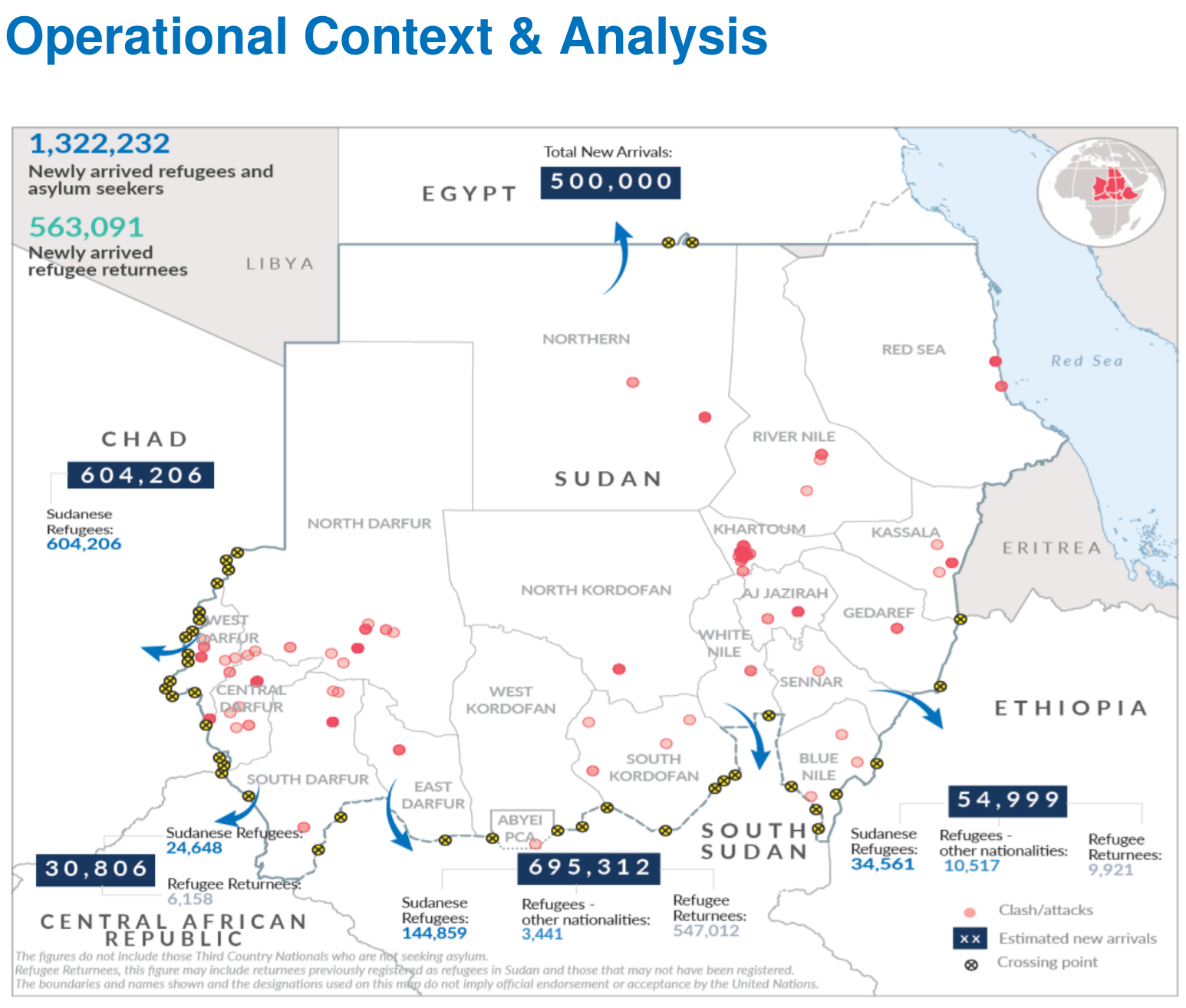}
        \caption{}
        \label{fig:unhcr-c}
    \end{subfigure}
    \hfill
    \begin{subfigure}[b]{0.48\textwidth}
        \centering
        \includegraphics[max width=\linewidth, max height=0.18\textheight, keepaspectratio]{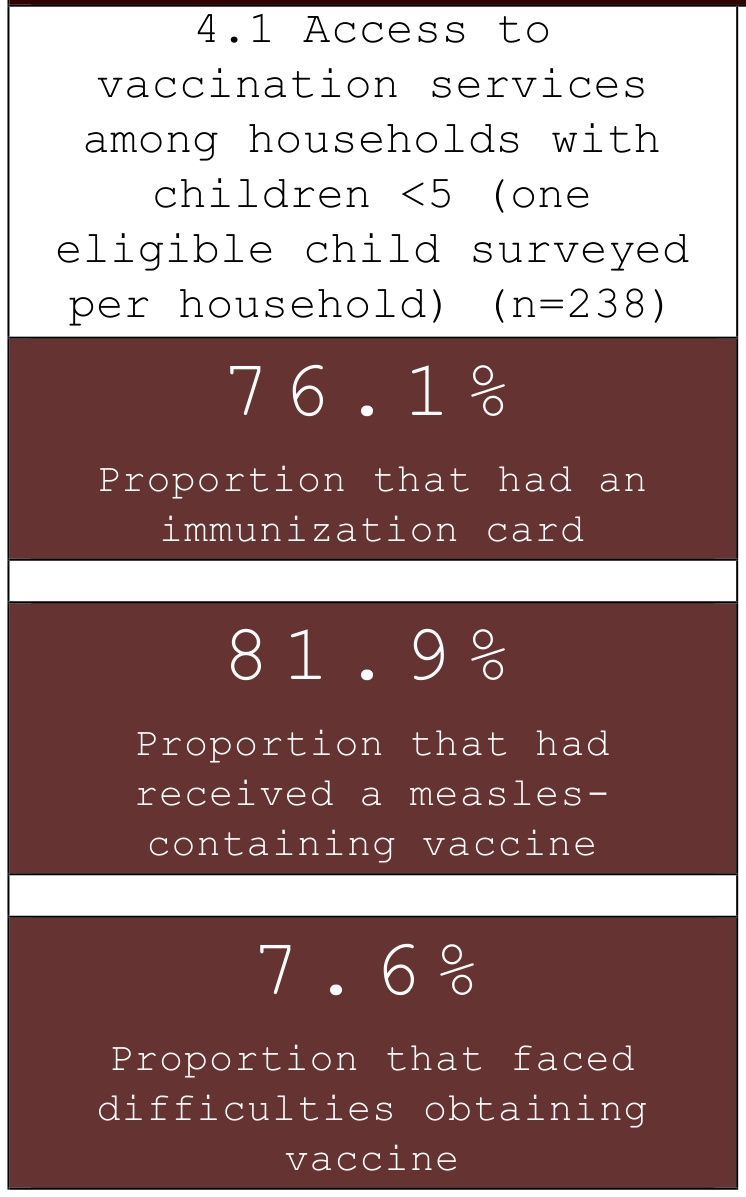}
        \caption{}
        \label{fig:unhcr-d}
    \end{subfigure}
    
    \vspace{0.5em}
    \begin{subfigure}[b]{0.48\textwidth}
        \centering
        \includegraphics[max width=\linewidth, max height=0.18\textheight, keepaspectratio]{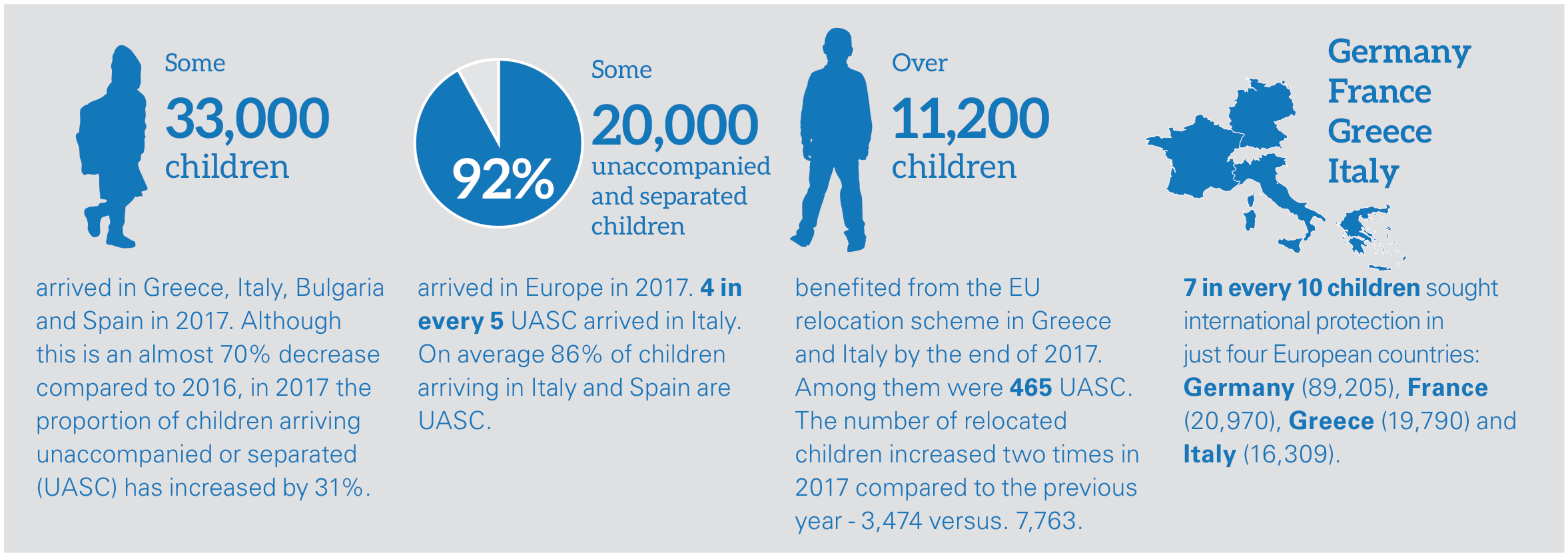}
        \caption{}
        \label{fig:unhcr-e}
    \end{subfigure}
    \hfill
    \begin{subfigure}[b]{0.48\textwidth}
        \centering
        \includegraphics[max width=\linewidth, max height=0.18\textheight, keepaspectratio]{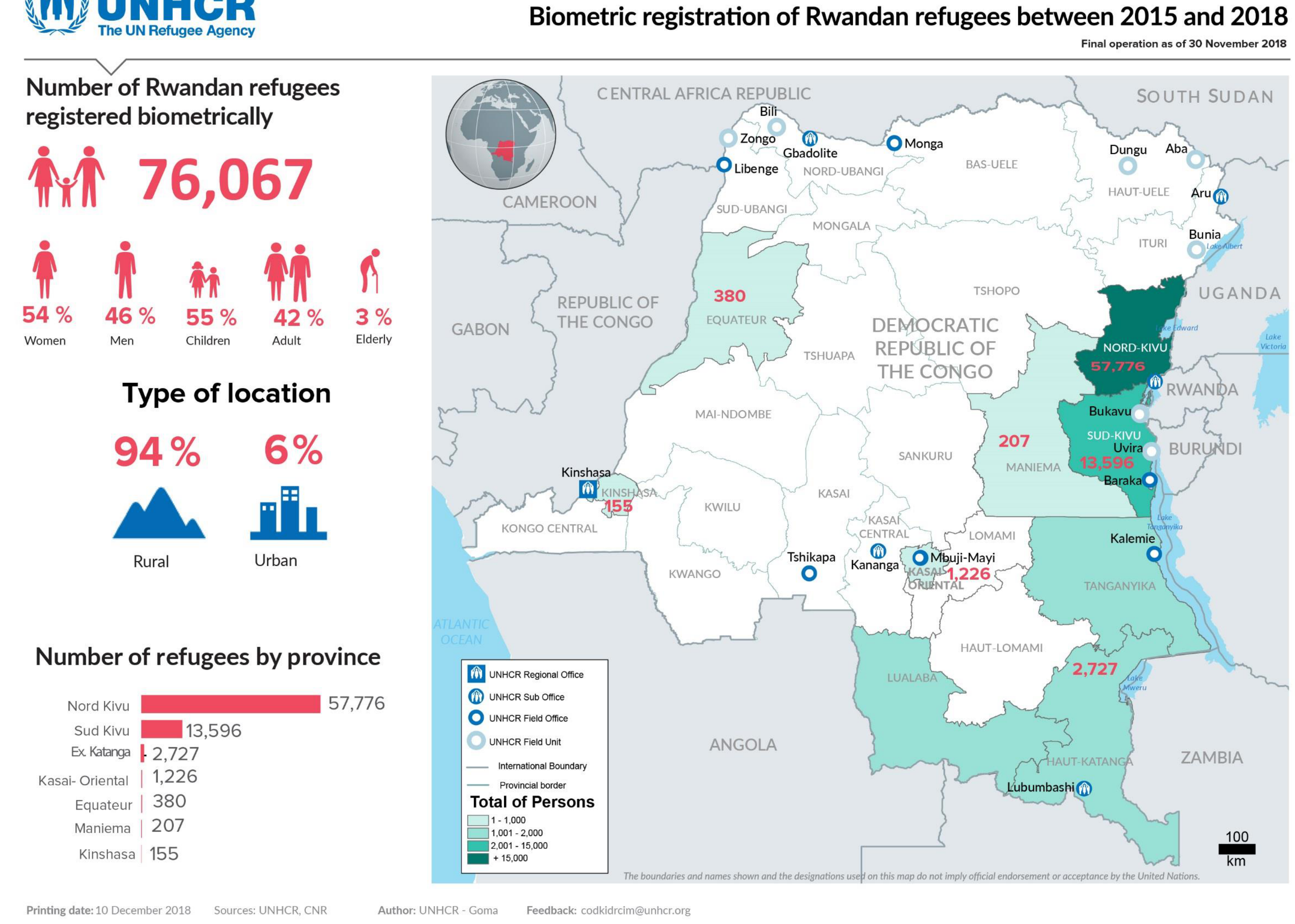}
        \caption{}
        \label{fig:unhcr-f}
    \end{subfigure}
    
    \vspace{0.5em}
    \begin{subfigure}[b]{0.48\textwidth}
        \centering
        \includegraphics[max width=\linewidth, max height=0.18\textheight, keepaspectratio]{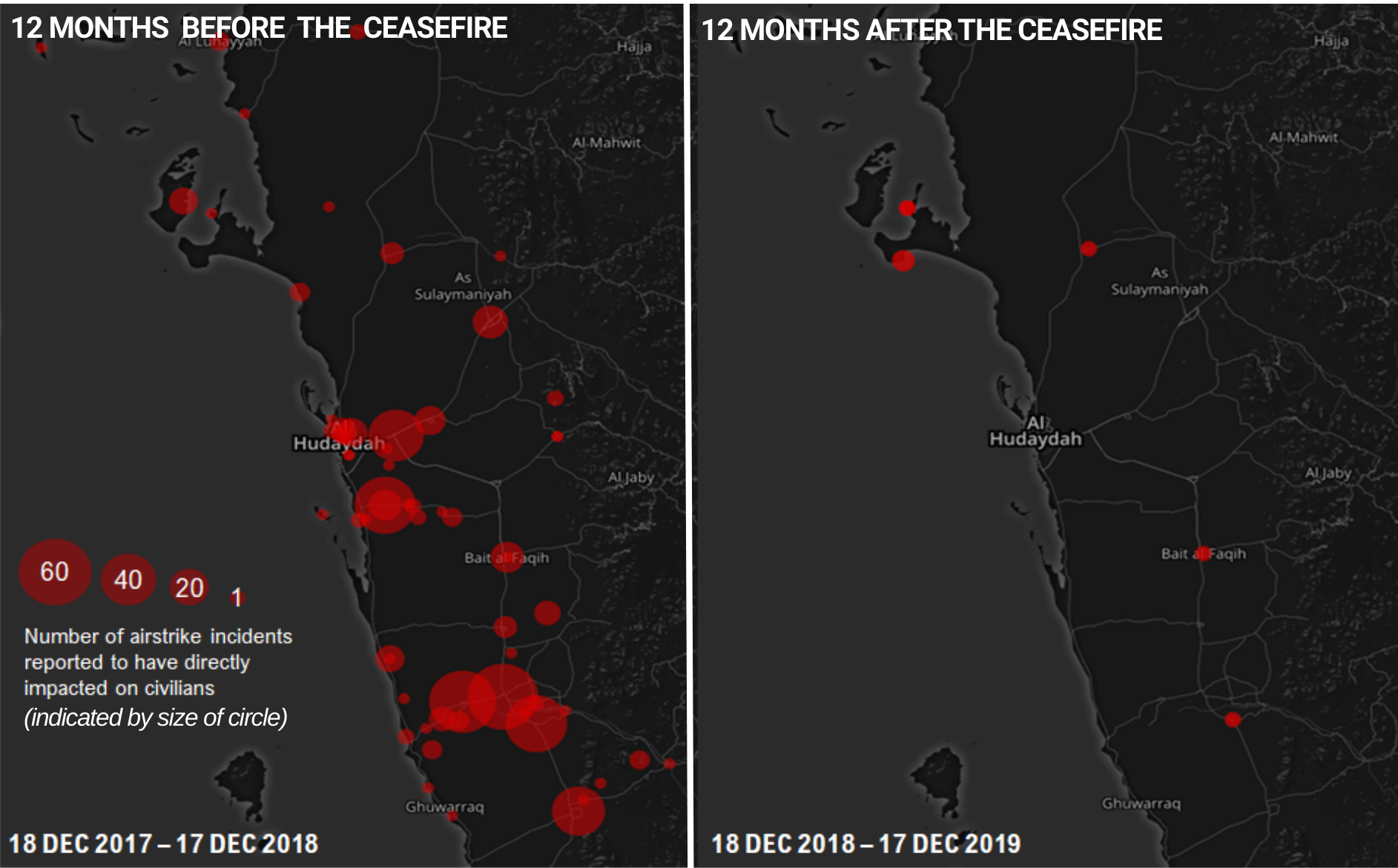}
        \caption{}
        \label{fig:unhcr-g}
    \end{subfigure}

    \caption{Representative data snapshots from the UNHCR / ReliefWeb corpus.
    (a) Multi-panel statistical summary.
    (b) Demographic infographic.
    (c) Operational monitoring map.
    (d) Statistical factoid panel.
    (e) Composite infographic.
    (f) Geospatial dashboard integrating maps and indicators.
    (g) Before--after spatial comparison.
    Together, these examples illustrate the diversity of analytical artifacts encountered in humanitarian reporting.}
    \label{fig:unhcr-examples}
\end{figure}

Figure~\ref{fig:unhcr-examples} presents representative examples from the corpus, including multi-panel statistical summaries (\ref{fig:unhcr-a}), infographic-style visualizations combining text and quantitative indicators (\ref{fig:unhcr-b}, \ref{fig:unhcr-e}), operational maps (\ref{fig:unhcr-c}), indicator panels (\ref{fig:unhcr-d}), composite dashboard-style monitoring summaries (\ref{fig:unhcr-f}), and comparative spatial analyses (\ref{fig:unhcr-g}). Together, these examples illustrate the wide range of visual forms through which operational information is communicated in humanitarian reporting.

A common characteristic across many examples is that their analytical meaning depends on the interaction between multiple constituent elements. Consequently, successful extraction requires identifying semantically coherent analytical artifacts rather than individual layout objects. This distinction is central to the data snapshot formulation adopted throughout the benchmark and helps explain why many existing layout detection systems struggle on humanitarian documents.

\subsection{Policy Research Working Papers (PRWP)}

The PRWP corpus is characterized by analytically dense tables and figures typical of academic economics, finance, and public policy research. Unlike humanitarian reporting, where information is often communicated through composite visual artifacts, PRWP snapshots primarily serve to present empirical evidence, statistical analyses, model estimates, and policy-relevant findings. Many snapshots contain information that is only partially discussed in the surrounding text, making them valuable standalone analytical artifacts.

\begin{figure}[htbp]
    \centering
    \begin{subfigure}[b]{0.48\textwidth}
        \centering
        \includegraphics[max width=\linewidth, max height=0.23\textheight, keepaspectratio]{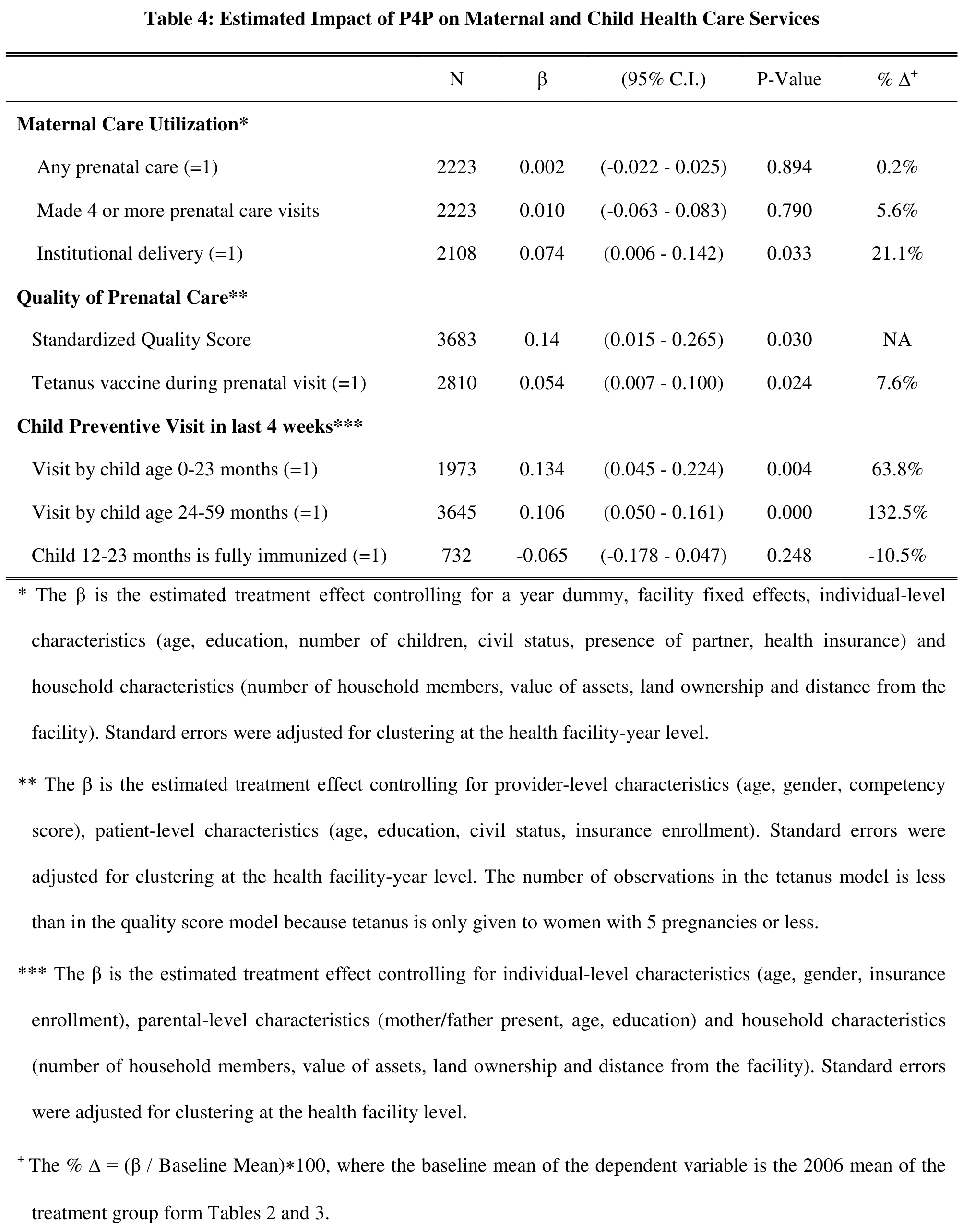}
        \caption{}
        \label{fig:prwp-a}
    \end{subfigure}
    \hfill
    \begin{subfigure}[b]{0.48\textwidth}
        \centering
        \includegraphics[max width=\linewidth, max height=0.23\textheight, keepaspectratio]{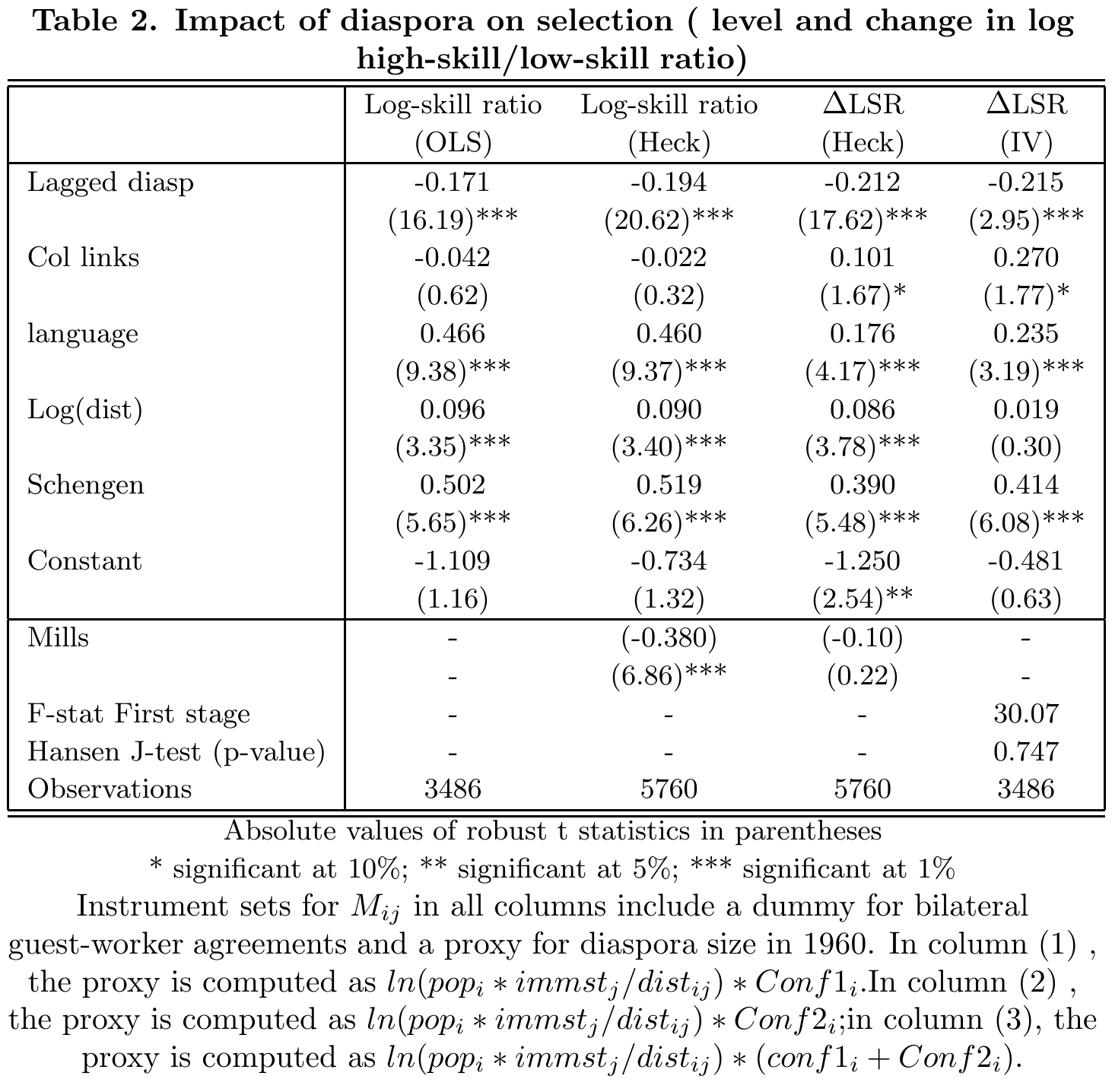}
        \caption{}
        \label{fig:prwp-b}
    \end{subfigure}
    
    \vspace{0.5em}
    \begin{subfigure}[b]{0.48\textwidth}
        \centering
        \includegraphics[max width=\linewidth, max height=0.23\textheight, keepaspectratio]{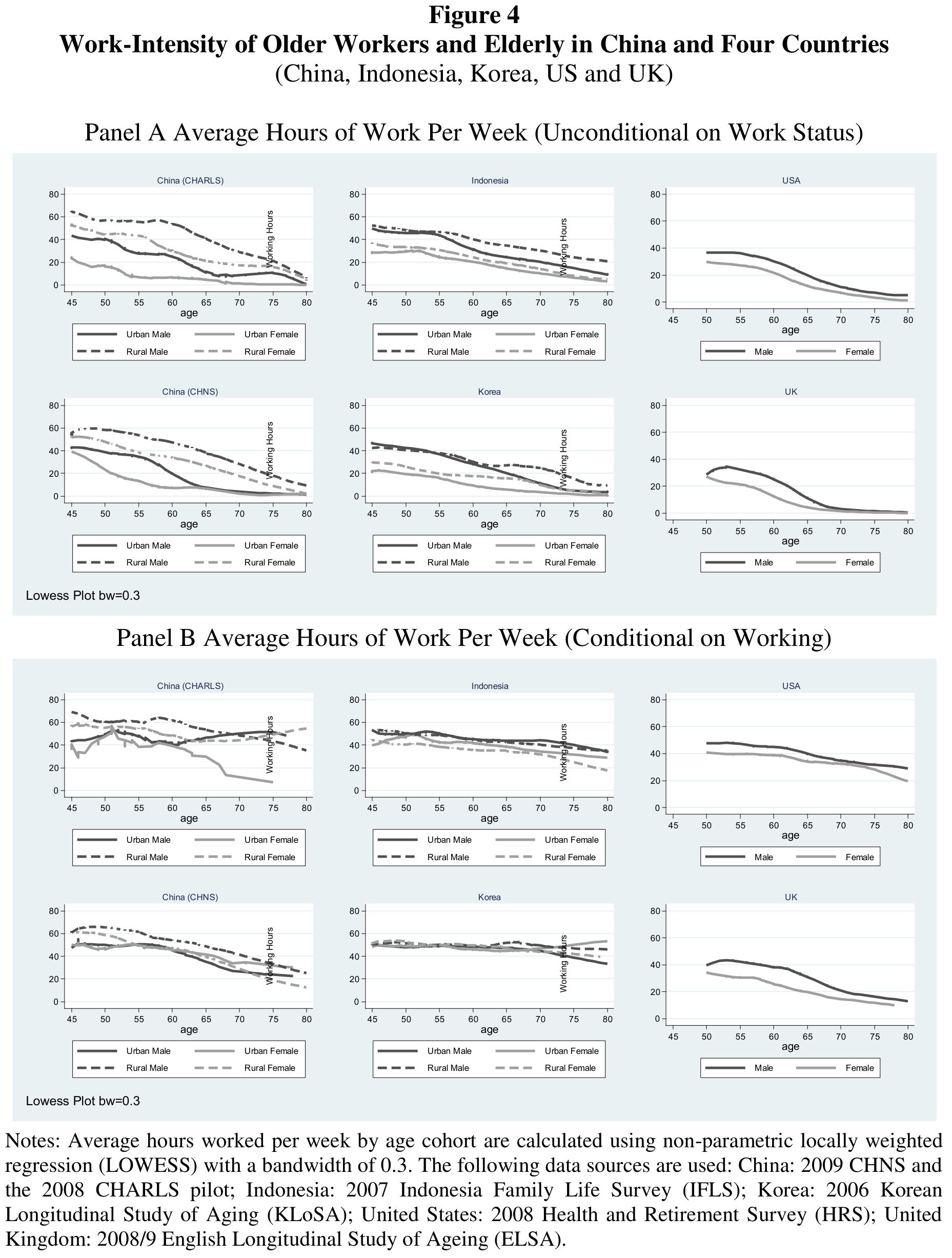}
        \caption{}
        \label{fig:prwp-c}
    \end{subfigure}
    \hfill
    \begin{subfigure}[b]{0.48\textwidth}
        \centering
        \includegraphics[max width=\linewidth, max height=0.23\textheight, keepaspectratio]{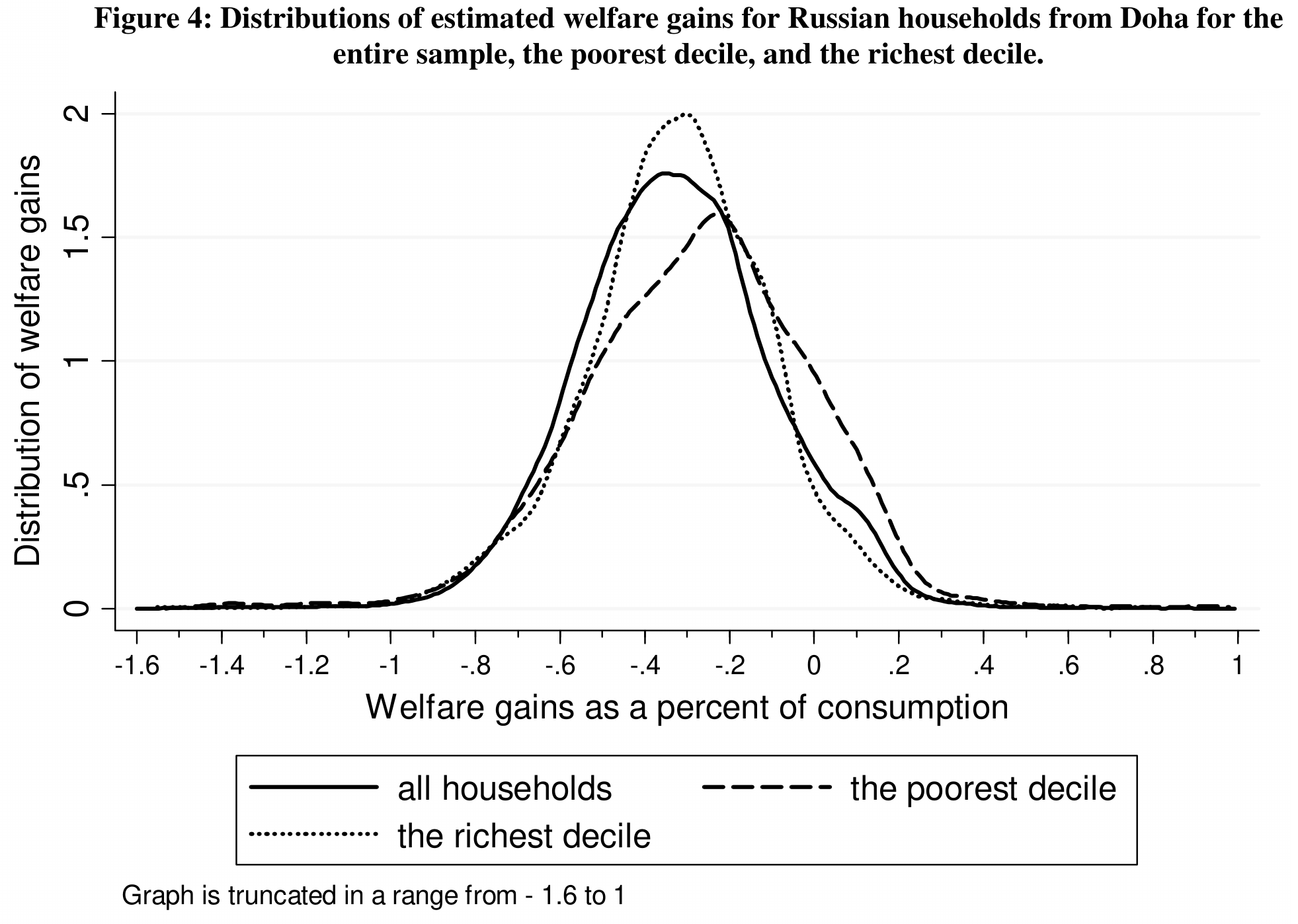}
        \caption{}
        \label{fig:prwp-d}
    \end{subfigure}
    
    \vspace{0.5em}
    \begin{subfigure}[b]{0.48\textwidth}
        \centering
        \includegraphics[max width=\linewidth, max height=0.23\textheight, keepaspectratio]{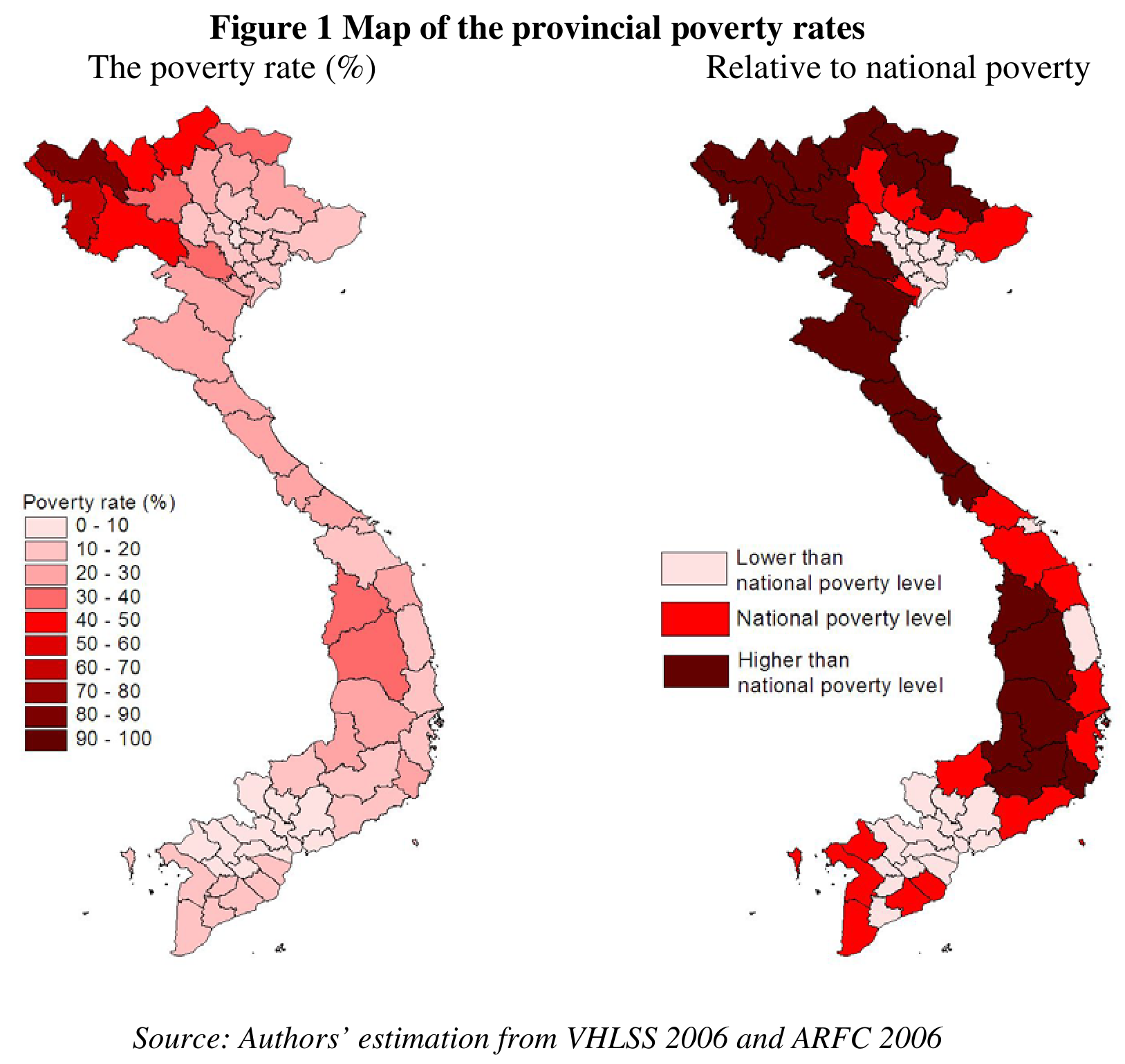}
        \caption{}
        \label{fig:prwp-e}
    \end{subfigure}
    \hfill
    \begin{subfigure}[b]{0.48\textwidth}
        \centering
        \includegraphics[max width=\linewidth, max height=0.23\textheight, keepaspectratio]{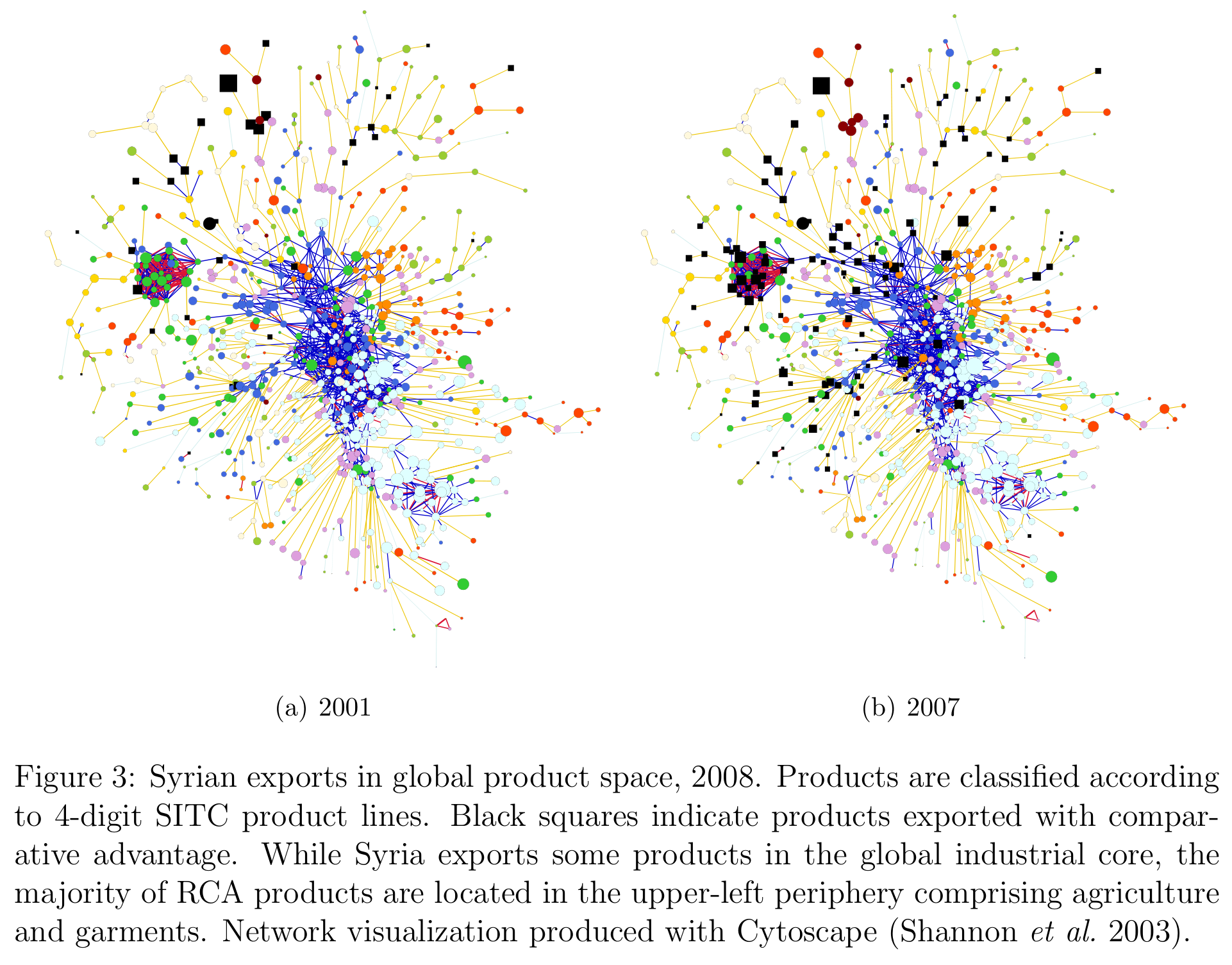}
        \caption{}
        \label{fig:prwp-f}
    \end{subfigure}

    \caption{Representative data snapshots from the PRWP corpus.
    (a) Impact evaluation table reporting estimated effects of a performance-based financing intervention on maternal and child health outcomes.
    (b) Econometric estimation table comparing alternative model specifications and identification strategies.
    (c) Multi-panel comparative analysis of labor-force participation among older workers across countries. 
    (d) Distributional analysis of estimated household welfare gains under a trade liberalization scenario.
    (e) Poverty mapping visualization showing regional variation in poverty rates across provinces.
    (f) Network representation of export structures within the global product space.
    Together, these examples illustrate the diversity of analytical artifacts commonly encountered in development economics and policy research.}
    \label{fig:prwp-examples}
\end{figure}

Figure~\ref{fig:prwp-examples} presents representative examples from the corpus, including an impact evaluation table reporting treatment effects across multiple health outcomes (\ref{fig:prwp-a}), an econometric estimation table comparing alternative model specifications and identification strategies (\ref{fig:prwp-b}), a multi-panel comparative analysis of labor-force participation across countries (\ref{fig:prwp-c}), a distributional analysis of estimated welfare gains (\ref{fig:prwp-d}), a poverty mapping visualization showing regional disparities (\ref{fig:prwp-e}), and a network representation of export structures within the global product space (\ref{fig:prwp-f}).

These examples illustrate the breadth of analytical artifacts encountered in development economics research. Although many conform to conventional notions of figures and tables, they frequently encapsulate the primary empirical findings of a study and often contain substantially more information than can be conveyed in the accompanying narrative. Consequently, successful extraction requires preserving complete analytical units—including titles, panel labels, legends, source information, and explanatory footnotes—so that the extracted snapshot remains interpretable in isolation. This characteristic motivates the benchmark's focus on semantically coherent data snapshots rather than individual document layout elements.

\subsection{Refugee PADs}

The Refugee PAD corpus differs from both the PRWP and UNHCR/ReliefWeb collections in that its visual artifacts are primarily designed to support project planning, implementation, monitoring, and resource allocation. Rather than presenting analytical findings or humanitarian situation updates, these documents emphasize operational information such as beneficiary targeting, financing structures, project indicators, implementation plans, and geographic distribution of services and populations. 

\begin{figure}[htbp]
    \centering
    \begin{subfigure}[b]{0.48\textwidth}
        \centering
        \includegraphics[max width=\linewidth, max height=0.23\textheight, keepaspectratio]{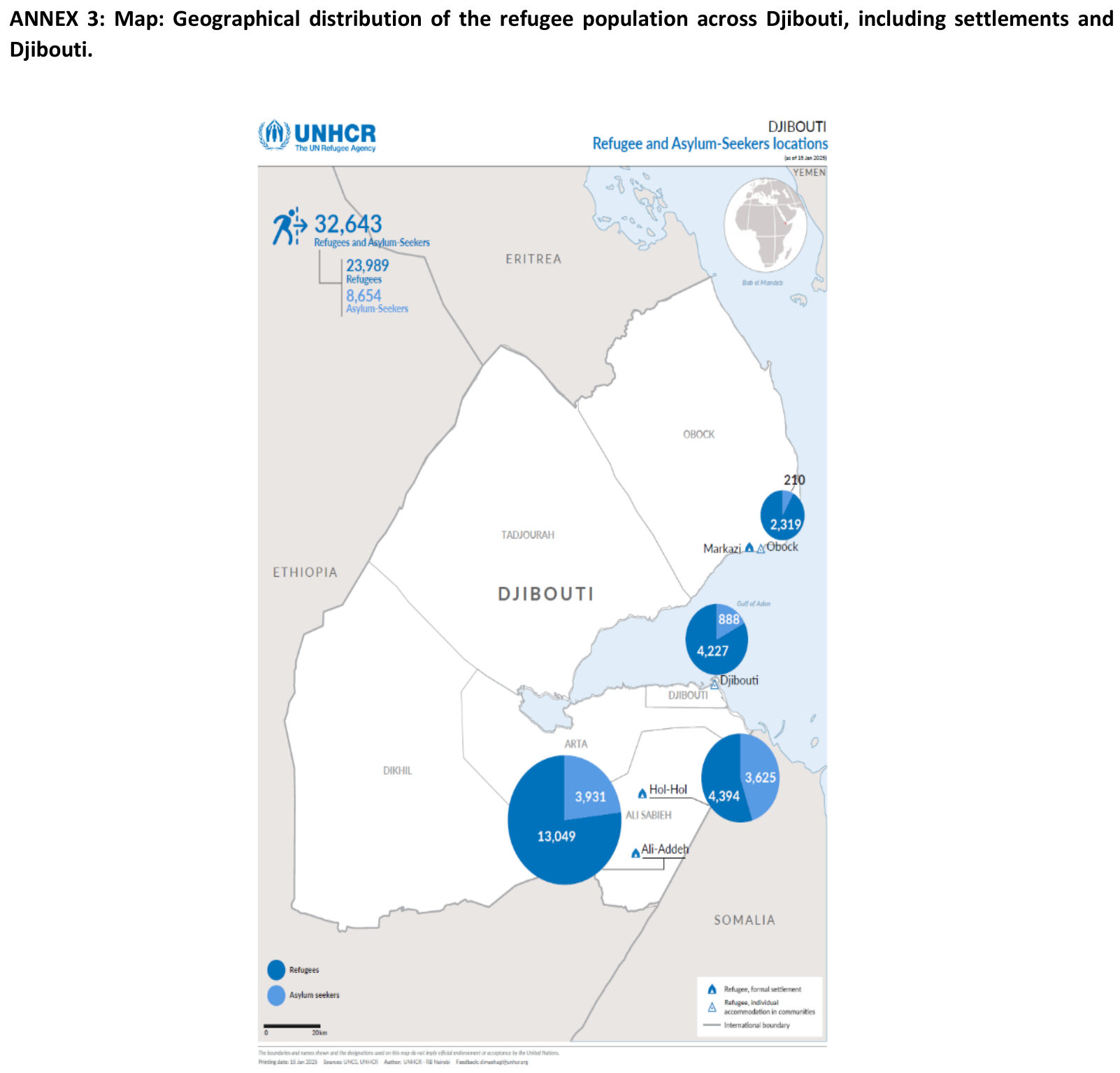}
        \caption{}
        \label{fig:refugee-a}
    \end{subfigure}
    \hfill
    \begin{subfigure}[b]{0.48\textwidth}
        \centering
        \includegraphics[max width=\linewidth, max height=0.23\textheight, keepaspectratio]{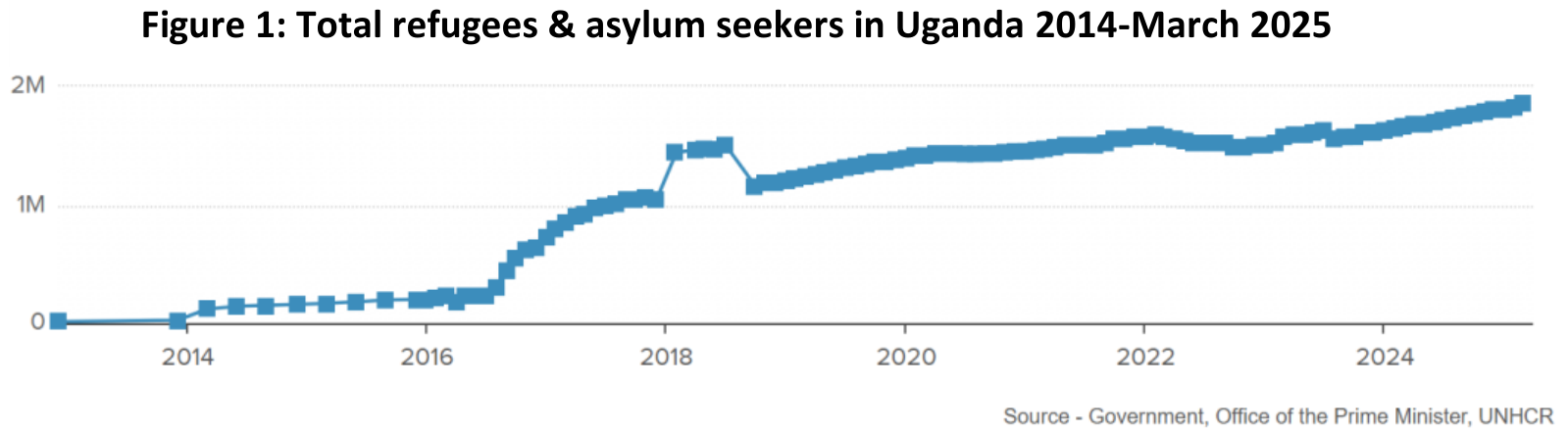}
        \caption{}
        \label{fig:refugee-b}
    \end{subfigure}
    
    \vspace{0.5em}
    \begin{subfigure}[b]{0.48\textwidth}
        \centering
        \includegraphics[max width=\linewidth, max height=0.23\textheight, keepaspectratio]{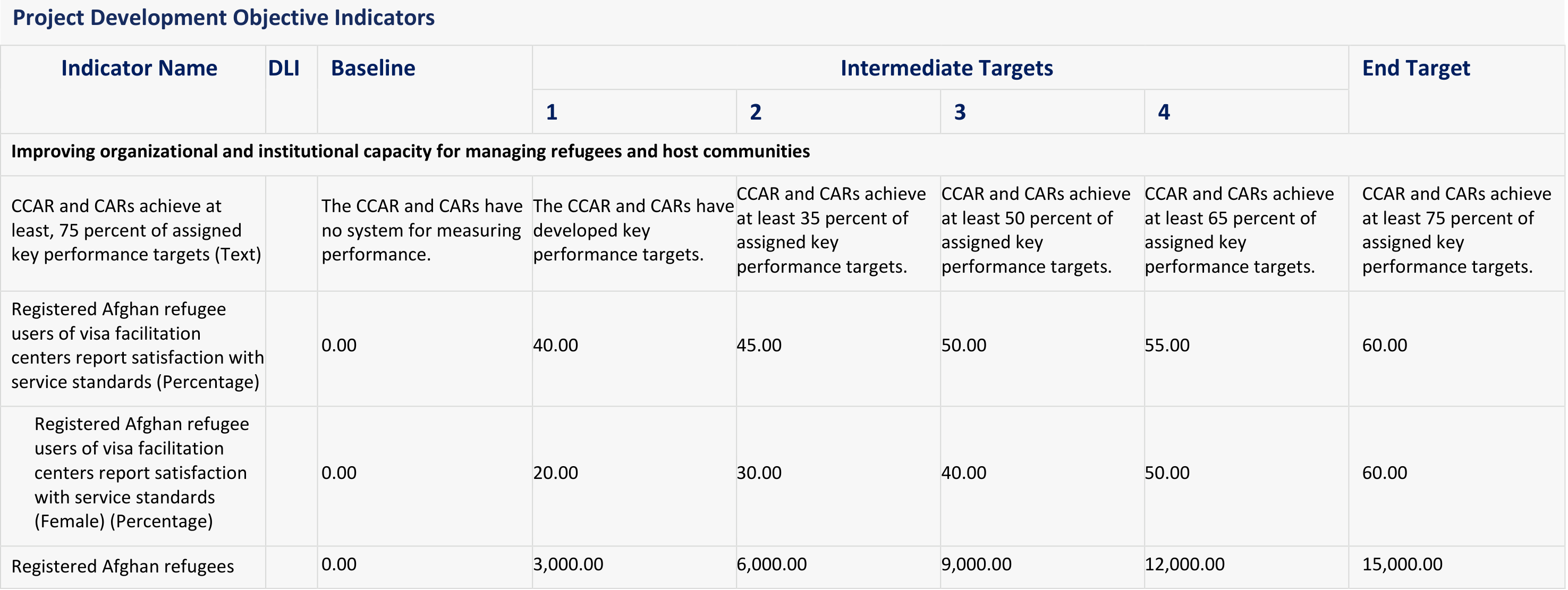}
        \caption{}
        \label{fig:refugee-c}
    \end{subfigure}
    \hfill
    \begin{subfigure}[b]{0.48\textwidth}
        \centering
        \includegraphics[max width=\linewidth, max height=0.23\textheight, keepaspectratio]{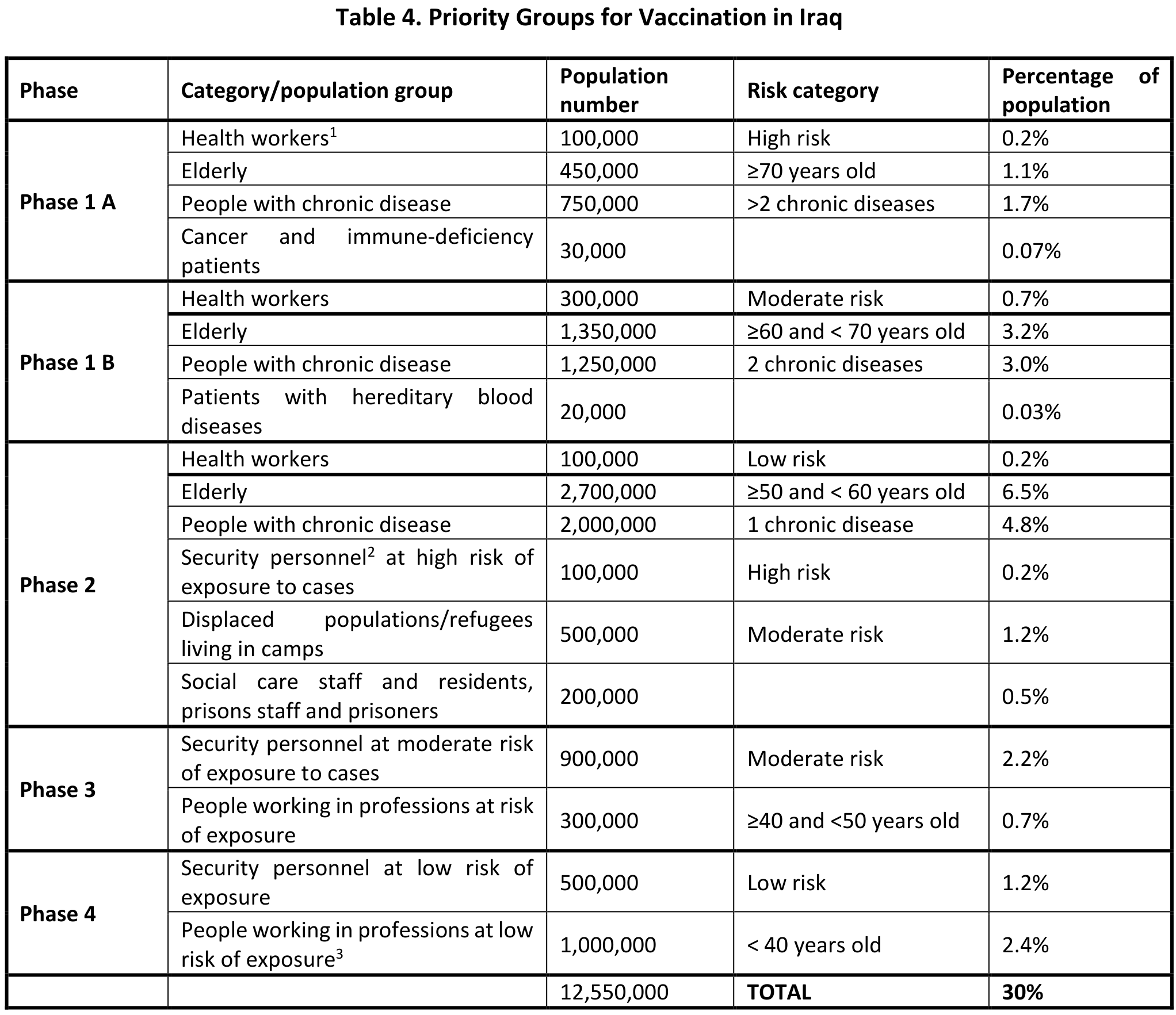}
        \caption{}
        \label{fig:refugee-d}
    \end{subfigure}
    
    \vspace{0.5em}
    \begin{subfigure}[b]{0.48\textwidth}
        \centering
        \includegraphics[max width=\linewidth, max height=0.23\textheight, keepaspectratio]{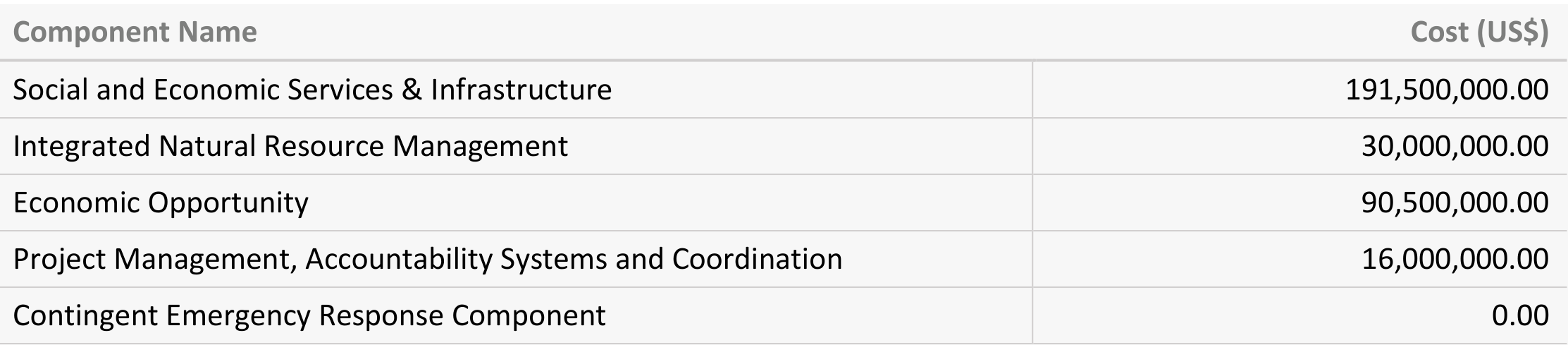}
        \caption{}
        \label{fig:refugee-e}
    \end{subfigure}
    \hfill
    \begin{subfigure}[b]{0.48\textwidth}
        \centering
        \includegraphics[max width=\linewidth, max height=0.23\textheight, keepaspectratio]{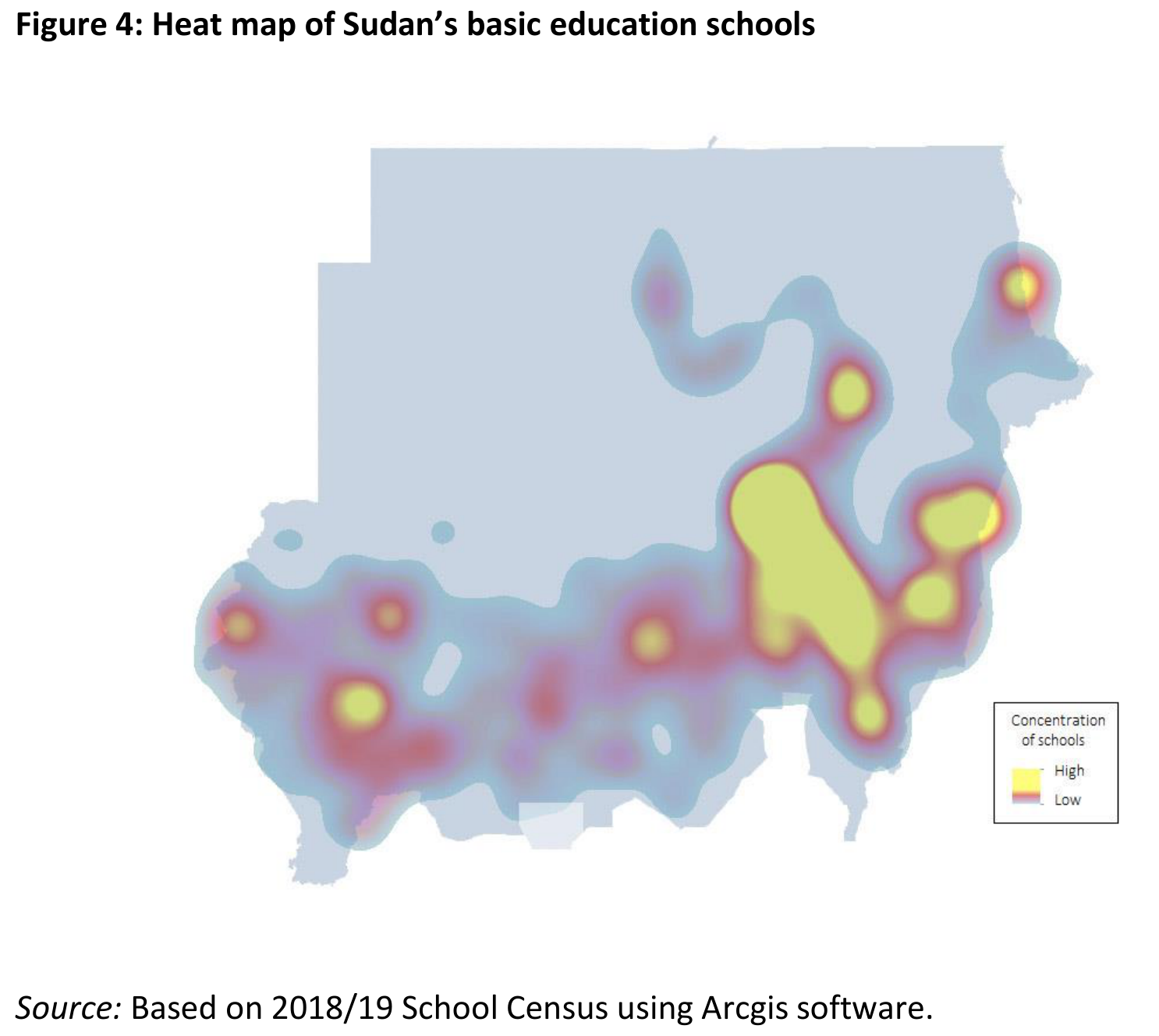}
        \caption{}
        \label{fig:refugee-f}
    \end{subfigure}

    \caption{Representative data snapshots from the Refugee PAD corpus.
    (a) Refugee and asylum-seeker distribution map.
    (b) Refugee population monitoring chart.
    (c) Project development objective indicator framework.
    (d) Vaccination prioritization and beneficiary targeting table.
    (e) Project component financing summary.
    (f) Spatial density visualization of service locations.
    Together, these examples illustrate the operational planning, monitoring, financing, and geographic information artifacts commonly encountered in refugee-focused project appraisal documents.}
    \label{fig:refugee-examples}
\end{figure}

Figure~\ref{fig:refugee-examples} presents representative examples from the corpus, including refugee distribution maps (\ref{fig:refugee-a}), temporal monitoring charts tracking refugee populations (\ref{fig:refugee-b}), results-framework indicator tables (\ref{fig:refugee-c}), beneficiary prioritization tables (\ref{fig:refugee-d}), project financing summaries (\ref{fig:refugee-e}), and spatial density visualizations (\ref{fig:refugee-f}). Together, these examples illustrate the diverse visual forms through which refugee-related development operations communicate planning assumptions, implementation targets, resource allocations, and monitoring information.

\begin{figure}[htbp]
    \centering
    \includegraphics[width=0.75\linewidth]{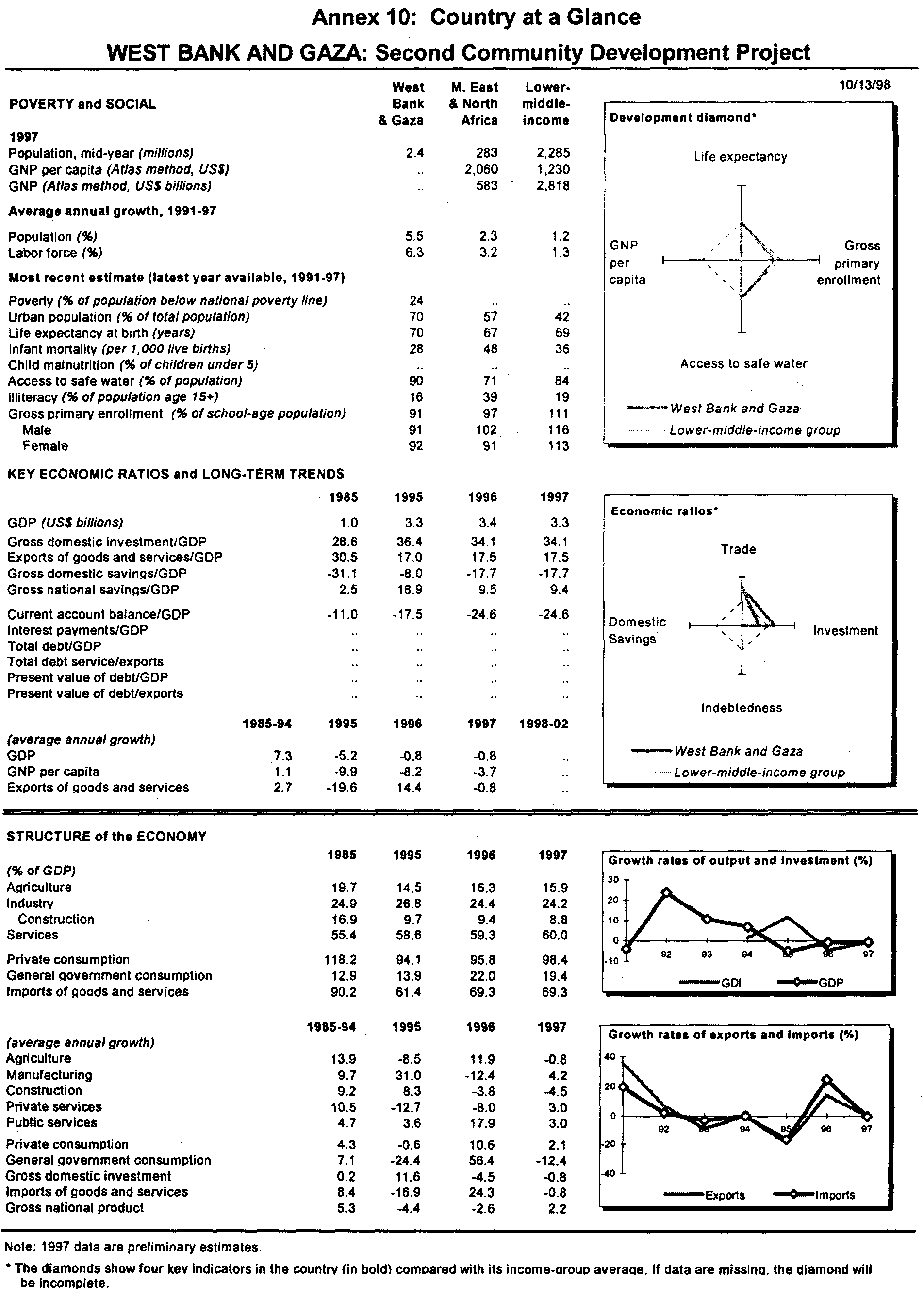}
    \caption{Example of a compound analytical dashboard from the Refugee PAD corpus. The page combines statistical tables, comparative indicators, radar charts, and time-series visualizations within a single integrated analytical artifact. Recovering the full analytical content requires preserving relationships across these heterogeneous components.}
    \label{fig:refugee-dashboard}
\end{figure}

A notable characteristic of the corpus is the prevalence of operational artifacts whose meaning depends on relationships spanning multiple rows, columns, legends, and visual elements. Results frameworks, financing summaries, and prioritization matrices encode hierarchical program structures that cannot be recovered from isolated table cells, while maps and spatial visualizations combine geographic context, quantitative indicators, and symbolic encodings within a single analytical artifact. The corpus also contains compound statistical dashboards (Figure~\ref{fig:refugee-dashboard}) that integrate multiple tables and visualizations on a single page, further illustrating the need to preserve relationships across heterogeneous components. Consequently, successful extraction requires recovering complete analytical artifacts rather than isolated layout elements, reinforcing the data snapshot formulation adopted in this benchmark.

\newpage

\section{Appendix: Non-rectangular data snapshots}
\label{sec:non-rectangular}

Data snapshots in the benchmark are represented using rectangular bounding boxes to maintain consistency with common document layout annotation practices. While this representation is appropriate for most figures and tables, some analytical artifacts cannot be cleanly represented using rectangular bounding boxes. In these cases, annotators must approximate the intended snapshot extent using the smallest bounding box that preserves the complete artifact.

\begin{figure}[htbp]
    \centering
    \begin{subfigure}[b]{0.48\textwidth}
        \centering
        \includegraphics[max width=\linewidth, max height=0.23\textheight, keepaspectratio]{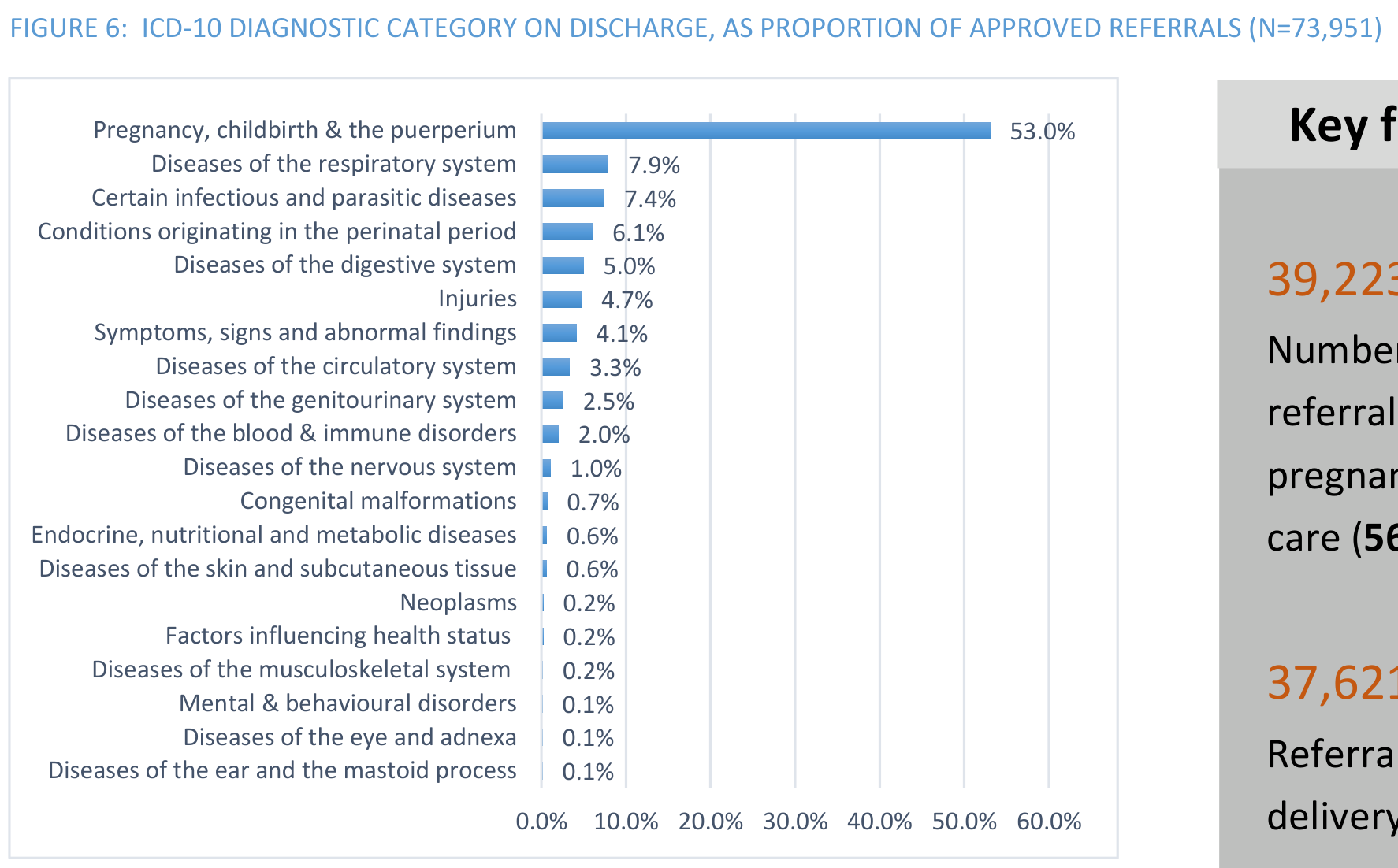}
        \caption{}
        \label{fig:nonrect-a}
    \end{subfigure}
    \hfill
    \begin{subfigure}[b]{0.48\textwidth}
        \centering
        \includegraphics[max width=\linewidth, max height=0.23\textheight, keepaspectratio]{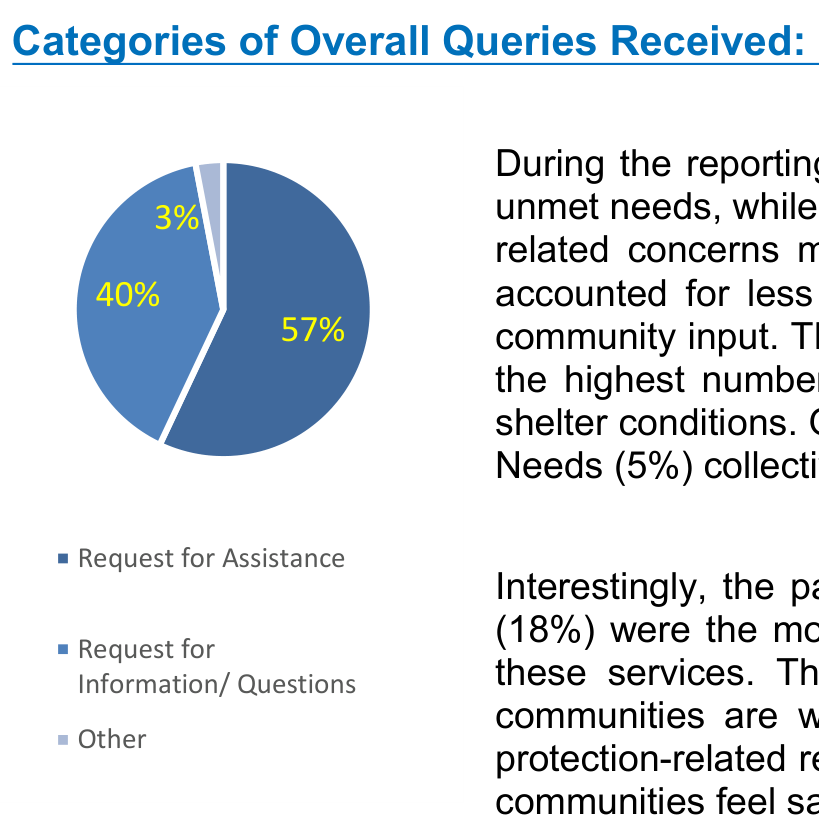}
        \caption{}
        \label{fig:nonrect-b}
    \end{subfigure}
    
    \vspace{0.5em}
    \begin{subfigure}[b]{0.48\textwidth}
        \centering
        \includegraphics[max width=\linewidth, max height=0.23\textheight, keepaspectratio]{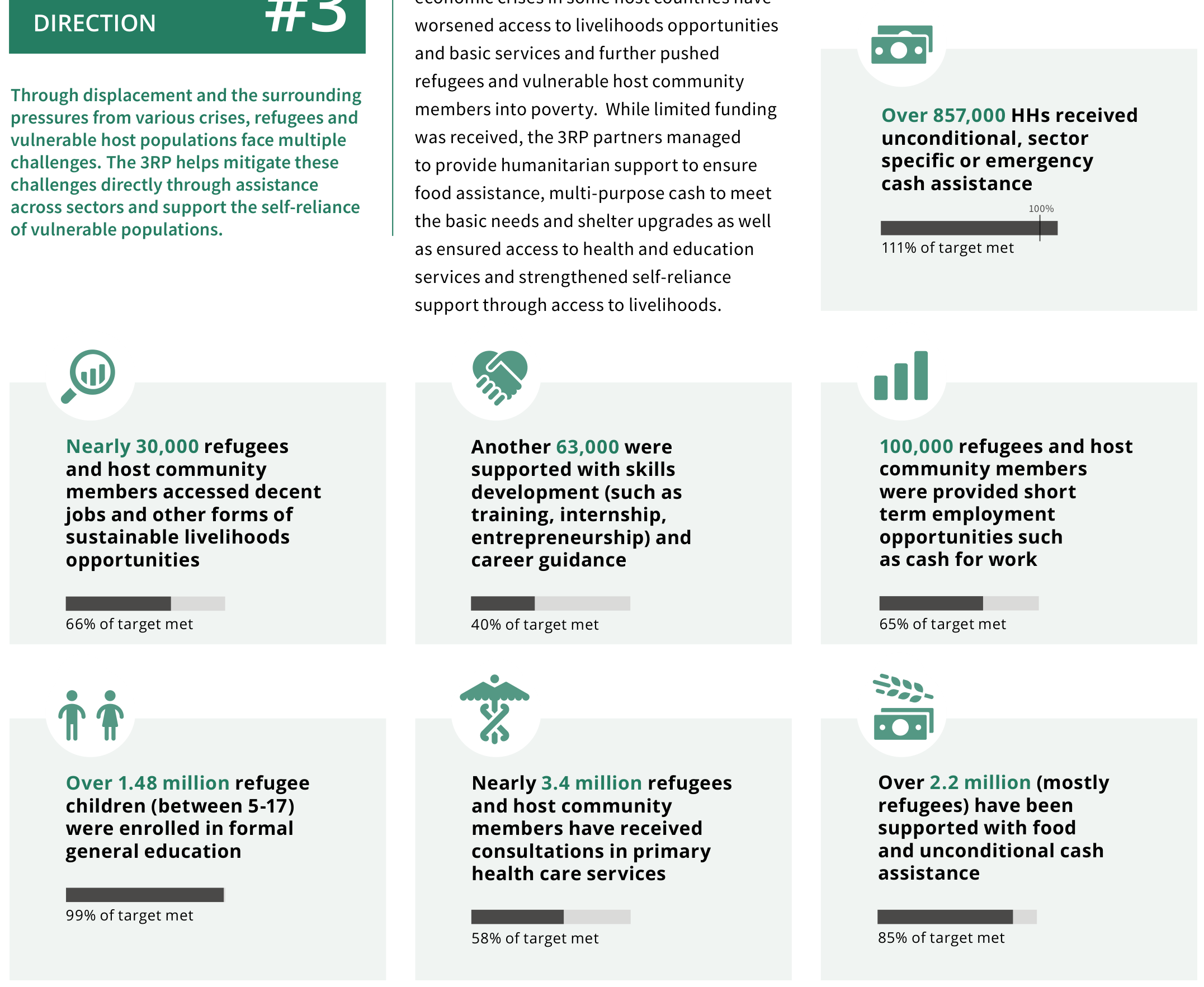}
        \caption{}
        \label{fig:nonrect-c}
    \end{subfigure}

    \caption{Examples illustrating limitations of rectangular bounding-box annotations. 
    (a) Statistical chart where preserving the complete title requires including part of a neighboring visual element.
    (b) Pie-chart visualization where preserving the title requires including adjacent narrative text that is not part of the intended data snapshot and may be mistaken for analytical content.
    (c) Multiple indicator panels that collectively form a single analytical artifact despite being spatially separated.
    These examples demonstrate how rectangular annotations may include limited amounts of surrounding content when approximating the extent of complex analytical artifacts.}
    \label{fig:nonrect}
\end{figure}

Figure~\ref{fig:nonrect} illustrates several representative examples. Neighboring visual elements may prevent a clean crop boundary without excluding information necessary for interpretation (\ref{fig:nonrect-a}). Preserving essential elements such as titles may also require the inclusion of adjacent narrative text that is not itself part of the analytical artifact and may be misinterpreted as analytical content by downstream systems (\ref{fig:nonrect-b}). Finally, a single snapshot may consist of multiple spatially separated components that must nevertheless be captured within a single bounding box (\ref{fig:nonrect-c}).

These examples highlight a practical limitation of rectangular annotations. Although bounding boxes provide a simple and scalable representation for benchmarking, they may introduce small amounts of irrelevant content, whitespace, or neighboring visual elements when used to approximate the extent of complex analytical artifacts.

\end{document}